\documentclass{article}

\usepackage[preprint]{neurips_2025}

\usepackage[utf8]{inputenc} % allow utf-8 input
\usepackage[T1]{fontenc}    % use 8-bit T1 fonts
\usepackage{hyperref}       % hyperlinks
\usepackage{url}            % simple URL typesetting
\usepackage{booktabs}       % professional-quality tables
\usepackage{amsfonts}       % blackboard math symbols
\usepackage{nicefrac}       % compact symbols for 1/2, etc.
\usepackage{microtype}      % microtypography
\usepackage{xcolor}         % colors
\usepackage{amsmath}
\usepackage{algorithm}
\usepackage{algpseudocode} 
\usepackage{amsthm}
\usepackage[pdftex]{graphicx}
\usepackage{wrapfig}

\newcommand{\logi}{\mathrm{LS}}

\theoremstyle{definition}
\newtheorem{definition}{Definition}[section]

\title{Geometric Embedding Alignment \\ via Curvature Matching in Transfer Learning}

\author{%
  Sung Moon Ko\thanks{Contributed equally} \\
  LG AI Research\\
  150, Magokjungang-ro, Gangseo-gu, Seoul, Korea Republic of \\
  \texttt{sungmoon.ko@lgresearch.ai} \\
  \And
  Jaewan Lee\footnotemark[1] \\
  LG AI Research\\
  \texttt{jaewan.lee@lgresearch.ai} \\
  \AND
  Sumin Lee\footnotemark[1] \\
  LG AI Research\\
  \texttt{sumin.lee@lgresearch.ai}  \\
  \And
  Soorin Yim\footnotemark[1] \\
  LG AI Research\\
  \texttt{soorin.yim@lgresearch.ai} \\
  \And
  Kyunghoon Bae \\
  LG AI Research\\
  \texttt{k.bae@lgresearch.ai}\\
  \And
  Sehui Han\\
  LG AI Research\\
  \texttt{hansse.han@lgresearch.ai}
}

\begin{document}

\maketitle

\begin{abstract}
  Geometrical interpretations of deep learning models offer insightful perspectives into their underlying mathematical structures. In this work, we introduce a novel approach that leverages differential geometry, particularly concepts from Riemannian geometry, to integrate multiple models into a unified transfer learning framework. By aligning the Ricci curvature of latent space of individual models, we construct an interrelated architecture, namely Geometric Embedding Alignment via cuRvature matching in transfer learning (GEAR), which ensures comprehensive geometric representation across datapoints. This framework enables the effective aggregation of knowledge from diverse sources, thereby improving performance on target tasks. We evaluate our model on 23 molecular task pairs sourced from various domains and demonstrate significant performance gains over existing benchmark model under both random (14.4\%) and scaffold (8.3\%) data splits.
\end{abstract}

%%%%%%%%%%%%%%%%%%%%%%%%%%%%%%%%%%%%%%%%%%%%%%%%%%%%%%%%%%%%%%%%%%%%%%%%%%
 %%%%%%%%%%%%%%%%%%%%%%%%%%%%%%%%%%%%%%%%%%%%%%%%%%%%%%%%%%%%%%%%%%%%%%%%%%
 %%%%%%%%%%%%%%%%%%%%%%%%%%%%%%%%%%%%%%%%%%%%%%%%%%%%%%%%%%%%%%%%%%%%%%%%%%
 
\section{Introduction}
Interest in the practical applications of deep learning has grown drastically over the years. Numerous examples have been announced recently, including applications in scientific domains such as biomedical, physical, and chemical sciences \citet{wang2019, doi:10.1073/pnas.2024383118, scarselli2009, bruna2013, duvenaud2015, defferrard2016, jin2018, C8SC04228D, ko2023grouping, ko2023geometricallyalignedtransferencoder, ko2024multitaskextensiongeometricallyaligned, yim2024taskadditionmultitasklearning, lee2024scalablemultitasktransferlearning}. However, in most real-world application cases—regardless of the domain—the lack of data consistently poses a major obstacle. Considerable efforts have been devoted to overcoming this challenge. One promising approach involves leveraging transfer learning (TL) and multitask learning (MTL) to make use of information across different datasets, modalities, and tasks. \citet{https://doi.org/10.1002/sam.10099, doi:10.1137/1.9781611972825.47, 10.1145/2433396.2433449, 6606822, 9051683, Quattoni, Kulis2011WhatYS, DBLP:journals/corr/abs-1902-07208, YU2022230}

TL, our primary focus, is a learning strategy that leverages information across different tasks to improve performance on a target task. Molecular property prediction tasks provide an excellent testbed for TL, as they typically involve relatively small datasets but a large number of prediction tasks per input molecule.

Most existing research has concentrated on classification tasks \citet{Radhakrishnan2023-tg, basu2023efficientequivarianttransferlearning, wenzel2022assayingoutofdistributiongeneralizationtransfer}, while relatively few approaches have been developed to support regression tasks—despite the fact that many practical applications in molecular sciences involve regression \citet{scarselli2009, bruna2013, duvenaud2015, defferrard2016, jin2018, C8SC04228D, ko2023grouping, ko2023geometricallyalignedtransferencoder, ko2024multitaskextensiongeometricallyaligned, yim2024taskadditionmultitasklearning, lee2024scalablemultitasktransferlearning}. Given the real-world importance of regression problems, this underrepresentation is notable. Therefore, in this work, we focus on the regression-based TL setting applied to molecular property prediction and propose a novel method specifically tailored to this context.

By analyzing the general structure of TL, one can observe that there is always a 'bridging' component that connects different tasks to facilitate the flow of information. Our method redefines and enhances this bridging mechanism by reinterpreting the latent space as a smooth, curved geometry. Since a key aspect of TL is designing effective methods to couple tasks, this geometric viewpoint allows us to align tasks by directly matching the geometric properties of their latent spaces.

\begin{figure}[ht]
\begin{center}
\includegraphics[trim={0cm 0cm 0cm 0cm}, clip, width=1\linewidth]{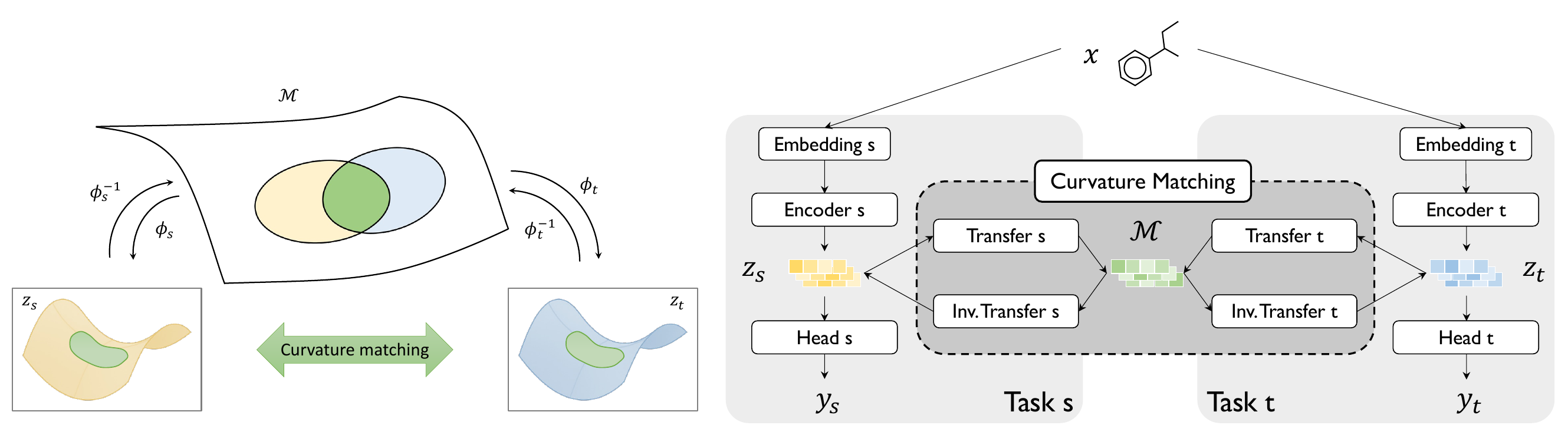}
\end{center}
\caption{(Left) The framework consists of a common manifold $\mathcal{M}$, task-specific latent spaces $z_s$ and $z_t$, transfer functions $\phi_s$ and $\phi_t$, and their inverses map $z_s$, $z_t$ and $\mathcal{M}$. (Right) Each task comprises five modules: embedding, encoder, transfer, inverse transfer, and head. Transfer and inverse transfer modules enable information exchange across tasks by curvature matching.}
\label{fig:algo_figure}
\end{figure}

The fundamental approach of our novel method is based on Riemannian differential geometry. This is a reasonable hypothesis, as most deep learning models are constructed using smooth functions to ensure the feasibility of backpropagation. Consequently, the latent spaces produced by these models can also be considered smooth, being composed through the application of these smooth functions. Several studies in this field \citet{ko2023grouping, ko2023geometricallyalignedtransferencoder, ko2024multitaskextensiongeometricallyaligned, yim2024taskadditionmultitasklearning, lee2024scalablemultitasktransferlearning} leverage the diffeomorphism invariance property of Riemannian geometry. These approaches have demonstrated effectiveness across multiple regression tasks in the molecular domain. However, despite their strengths, they also exhibit three major limitations inherent to their core algorithms.

One key limitation lies in the geometrical coverage of these algorithms. Because they operate by aligning infinitesimal distances between local perturbations, their effectiveness is inherently confined to local regions of the latent space. As a result, they struggle to capture the global geometric structure of the latent manifold. Furthermore, improving coverage typically requires increasing the number of perturbation points, which in turn leads to significant computational overhead.

Another important limitation is the potentially improper definition of `infinitesimal.' In cases where the latent space of a task exhibits high curvature or warping, some perturbation points may no longer be validly considered infinitesimal. This undermines the core assumption of local linearity and can lead to inaccurate geometric alignment.

The final limitation lies in the necessity of a shared embedding layer across all tasks. To align infinitesimal distances between tasks, the perturbation points must be consistently defined within a shared latent space, which in turn requires a shared embedding layer. However, such a mechanism is often inadequate for handling inputs associated with different levels of complexity.

Hence, hereby, we propose a model structure named the Geometric Embedding Alignment via cuRvature matching in TL (GEAR). Unlike previous approaches that rely on the diffeomorphism trick, our algorithm is built upon direct curvature matching, which, in turn, relaxes constraints on the input embedding structure. This allows for greater flexibility in customizing the model for individual tasks. We conducted regression experiments comparing GEAR to conventional TL methods using 23 pairs of molecular properties collected from various databases, and demonstrated that GEAR significantly outperforms existing approaches in most test cases. Furthermore, we validated the model’s robustness through a series of ablation studies.

Our main contribution of the article is as follows.
\begin{itemize}
    \item We design a novel TL algorithm GEAR based on Ricci curvature matching of latent spaces.
    \item GEAR significantly outperforms benchmark models in various molecular property regression tasks.
    \item GEAR exhibits stable geometry and robust behavior in extrapolation tasks.
\end{itemize}
%%%%%%%%%%%%%%%%%%%%%%%%%%%%%%%%%%%%%%%%%%%%%%%%%%%%%%%%%%%%%%%%%%%%%%%%%%
 %%%%%%%%%%%%%%%%%%%%%%%%%%%%%%%%%%%%%%%%%%%%%%%%%%%%%%%%%%%%%%%%%%%%%%%%%%
 %%%%%%%%%%%%%%%%%%%%%%%%%%%%%%%%%%%%%%%%%%%%%%%%%%%%%%%%%%%%%%%%%%%%%%%%%%

\section{Related works}
\label{Related Works}
\subsection{Riemannian geometry in deep learning}

Geometric deep learning is a field that extends deep learning to non-Euclidean domains such as graphs and manifolds, gaining prominence for its ability to capture complex relational and structural patterns inherent in scientific, biological and real-world data\citet{bronstein2017geometric}. Riemannian differential geometry is a branch of mathematics that studies smooth manifolds equipped with a metric that allows the measurement of lengths and angles on the manifold. In the context of deep learning, this framework is instrumental in understanding and modeling the geometric structure of data, particularly in high-dimensional spaces. By treating data as lying on a manifold, Riemannian geometry facilitates the development of algorithms that respect the intrinsic geometry of the data, leading to more meaningful representations and improved performance in tasks such as classification\citet{pegios2024counterfactual, lee2022statistical}, clustering\citet{hu2024enhanced, yang2018geodesic}, and generative modeling\citet{park2023understanding, grattarola2019adversarial}. Riemannian metric learning enhances deep learning by enabling models to operate in geometrically meaningful ways, improving interpretability and performance beyond Euclidean limits\citet{li2023geometry, sun2024geometry}.

\subsection{Transfer learning for molecular property prediction}
TL has shown significant promise in molecular property prediction, particularly in data-scarce settings. \citet{falk2023transfer} combine graph neural networks (GNNs) with kernel mean embeddings to enable knowledge transfer across atomistic simulations, capturing both local and global chemical features. \citet{buterez2024transfer, hoffmann2023transfer} further extend this by leveraging multi-fidelity datasets, demonstrating that pretraining on low-fidelity data and fine-tuning on high-fidelity targets significantly improves molecular property prediction. \citet{yao2024fast} quantify task relatedness between molecular property prediction datasets, providing guidance for effective TL to enhance prediction performance.

In addition, recent studies have begun incorporating Riemannian differential geometry into TL frameworks for molecular property prediction. In \citet{ko2023geometricallyalignedtransferencoder}, source and target tasks are aligned by matching distances in infinitesimal regions of the latent space. The method is later generalized to a multi-task setup involving more than two tasks in \citet{ko2024multitaskextensiongeometricallyaligned}. However, due to the computational burden of scaling this approach to many tasks, \citet{yim2024taskadditionmultitasklearning} introduce a task addition strategy to accelerate training.
%%%%%%%%%%%%%%%%%%%%%%%%%%%%%%%%%%%%%%%%%%%%%%%%%%%%%%%%%%%%%%%%%%%%%%%%%%
 %%%%%%%%%%%%%%%%%%%%%%%%%%%%%%%%%%%%%%%%%%%%%%%%%%%%%%%%%%%%%%%%%%%%%%%%%%
 %%%%%%%%%%%%%%%%%%%%%%%%%%%%%%%%%%%%%%%%%%%%%%%%%%%%%%%%%%%%%%%%%%%%%%%%%%

\section{Methods}
\label{Methods}
A geometric interpretation of latent space requires some mathematical preliminaries. The appropriate mathematical framework for describing curved spaces is differential geometry. Therefore, we briefly introduce the fundamental definitions and expressions that will be used in the following sections, along with the core ideas underlying our proposed method.

Since deep learning models always have a smooth underlying structure due to the backpropagation algorithm, it is very natural to assume that the latent space forged by a model is also smooth. Hence, it is plausible to assume the space is Riemannian. 

Let us consider the space in which the input dataset resides. This space contains all the information that can be utilized to perform any kind of downstream task. When a specific downstream task is fixed, the latent space formed by the downstream model effectively retracts the original space into a smaller, task-specific subspace to enhance performance. However, since the latent vectors originate from the same universal input space, the latent vector corresponding to a different downstream task should also represent the same point in that universal space. To reconcile latent representations from different downstream models, we leverage diffeomorphism invariance to construct an intermediate space with a locally flat frame, allowing us to align latent vectors from distinct downstream tasks.

Now, the real question is: how? In previously published methods~\citet{ko2023geometricallyalignedtransferencoder, ko2024multitaskextensiongeometricallyaligned, yim2024taskadditionmultitasklearning}, a perturbation strategy is used to align task-specific spaces. However, this approach has several notable drawbacks such as limited coverage of geometries and the requirement of a shared embedding layer. To address these issues, we extend the underlying idea by aligning the geometries of latent spaces through the matching of Ricci curvatures computed from each space. Since the computation of the Ricci scalar is highly intricate, we first introduce the basic forms of its constituent elements here, and provide a more detailed theoretical and mathematical walkthrough in the Appendix~\ref{proofs} and \ref{curvature computation}.

\subsection{Preliminary}\label{preliminary}
To maintain abstract notation, we will use the Einstein summation convention with index contraction representation. A fundamental introduction to these concepts is provided in the Appendix~\ref{notations}.

Riemannian geometry is often characterized by the Ricci scalar curvature. To understand how curvature is induced, one must carefully follow a step-by-step calculation process.

Everything begins with the metric. A metric is a rank-2 tensor with a symmetric property, which is crucial for computing distances between two points on a curved space. However, there is no systematic method to derive the explicit form of the metric for a given space directly. Instead, one must rely on a key mathematical property of Riemannian manifolds.

A Riemannian manifold always guarantees diffeomorphism invariance—in other words, freedom in the choice of coordinates on the manifold. This allows for the existence of a locally flat coordinate system under any circumstance. In such a system, the metric can be induced from the flat metric $\eta_{ij}$ by applying the Jacobian of the coordinate transformation at a given point. Here, $x'^i$ and $x^i$ are points on curved and locally flat frame respectively, and then, the Jacobian of the transformation between these coordinates then takes the following form.
\begin{equation}
    J^{i}_{\phantom{i}j} = \frac{dx'^i}{dx^j}
\end{equation}
From the Jacobian $J^i_{\phantom{i}j}$, one can compute the induced metric in a straightforward manner.
\begin{equation}\label{indmet}
    g_{ij} = \frac{dx'^m}{dx^i} \frac{dx'^n}{dx^j} \eta_{mn} = \frac{dx'^m}{dx^i} \frac{dx'_m}{dx^j}
\end{equation}
By obtaining the curved metric $g_{ij}$, one can define the Christoffel symbols $\Gamma^i_{\phantom{i}jk}$, which are used to construct the covariant derivative $\nabla_i$ —replacing the ordinary derivative in Riemannian geometry.
\begin{equation}\label{affine_connection}
    \Gamma^i_{\phantom{i}jk} = \frac{1}{2}g^{im}(\partial_j g_{mk} + \partial_k g_{mj} - \partial_m g_{kj})
\end{equation}
And the covariant derivative takes the following form.
\begin{equation} \label{cov_dev}
    \nabla_j T^i = \partial_j T^i + \Gamma^i_{\phantom{i}j l}T^l
\end{equation}
The curvature of a manifold $R^i_{\phantom{i}ljk}$ can be defined by the difference between tangent vectors that are parallel transported along different paths from the same initial point to the same final point. This can be expressed using the commutation relation of two covariant derivatives acting on a tangent vector.
\begin{equation} \label{Riemann_curv}
    R^{i}_{\phantom{i}ljk}T^{l} = [\nabla_j, \nabla_k] T^i
\end{equation}
Finally, by contracting $i$ and $j$, and $l$ and $k$ respectively, the Ricci scalar curvature $R$ can be obtained.
\begin{equation} \label{scalar_curv}
    R = g^{ij} g^{lk} R_{iljk}
\end{equation}
The scalar curvature is invariant under diffeomorphisms, as is evident from the fact that it has no free indices. Consequently, this quantity is often used to characterize the curvature of a given manifold. Since we are working with curved latent spaces and aiming to connect two different curved coordinate representations originating from a universal curved manifold, we directly compute this scalar property and align it to match the shapes of the task-specific spaces.

\subsection{Analytic computation strategy}\label{strategy}
A deep learning model is composed of multiple smooth layers. Therefore, if differentiable activation functions are used, it becomes possible to compute the curvature tensor of the curved space induced by the model. However, when the model consists of many layers, it becomes convenient to define building blocks that allow the full Jacobian to be computed by simply multiplying them. These building blocks can be expressed in terms of the weights and biases of each layer. Starting from the full Jacobian, and by applying the chain rule, the Jacobian can be decomposed into the Jacobians of individual layers.
\begin{equation}
    J^{i}_{\phantom{i}j} = \frac{dx^{'i}}{dx^{j}} = \frac{dx^{'i}}{dx^{(n-1)k_{n-1}}} \frac{dx^{(n-1)k_{n-1}}}{dx^{(n-2)k_{n-2}}} \cdots \frac{dx^{(1)k_1}}{dx^j}
\end{equation}
Here, $n$ denotes the layer index of the transfer module in the model, as illustrated in Figure~\ref{fig:algo_figure}. Therefore, when similar mathematical structures appear across layers—as is often the case—it becomes possible to define a fundamental building block of the full Jacobian using the Jacobian of a single representative layer. In our setup, each layer follows a linear MLP structure with the SiLU activation function. The fundamental Jacobian block can then be expressed in the following form:
\begin{eqnarray} \label{fundjaco}
\begin{array}{rl}
\frac{dx^{(n+1)i}}{dx^{(n)j}} =& W^{(n+1)i}_{\phantom{(n+1)m}k} (((x^{(n)k})e^{-x^{(n)k}} \times \logi(x^{(n)k}) + 1)\logi(x^{(n)k}))^k_{\phantom{k}j} \\
    =& (W^{(n+1)i}_{\phantom{(n+1)i}j}\sigma^{i} + (W^{(n+1)}x^{(n)}+b^{(n+1)})^{i}W^{(n+1)a_3}_{\phantom{(n+1)a_3}j}E^{i}_{\phantom{i}a_3}(\sigma^2)^{i})
\end{array}
\end{eqnarray}
Here, $W^{(n)i}_{\phantom{(n)i}j}$ and $b^{(n)i}$ are weights and biases of $n$-th layer in the transfer module. The new notations introduced in the equation above are defined as follows. First, $\textrm{LS}(x)$ denotes the logistic function and $\sigma^i$ and $E^i_{\phantom{i}l}$ are expressed as follows:
% \begin{equation}
%     \logi(x) \equiv \frac{1}{1+e^{-x}}
% \end{equation}
\begin{equation}
    \sigma^i = \frac{1}{1+e^{-(W^{(1)i}_{\phantom{(1)i}j} x^j + b^{(1)i})}}, \qquad \textrm{E}^{i}_{\phantom{i}l} \equiv (e^{-(W^l_{\phantom{l}j}x^j + b^j)})^{i}_{\phantom{i}l} = \Bigg\{
\begin{array}{cl}
     e^{-(W^l_{\phantom{l}j}x^j + b^l)} & \textrm{if $l = i$}\\
      0   & \textrm{if $l \neq i$}
\end{array}
\end{equation}
$(\sigma^2)^i$ denotes the element-wise square of $\sigma^i$.
% \begin{equation}
%     \textrm{E}^{i}_{\phantom{i}l} \equiv (e^{-(W^l_{\phantom{l}j}x^j + b^j)})^{i}_{\phantom{i}l} = \Bigg\{
% \begin{array}{cl}
%      e^{-(W^l_{\phantom{l}j}x^j + b^l)} & \textrm{if $l = i$}\\
%       0   & \textrm{if $l \neq i$}
% \end{array}
% \end{equation}
By utilizing Eq.~\ref{fundjaco}, it is now possible to compute the full Jacobian of the transfer module. The induced metric can then also be specified by Eq.~\ref{indmet}.

%%%%%%%%%%%%%%%%%%%%%%%%%%%%%%%%%%%%%%%%%%%
%여기서부터~~~
%%%%%%%%%%%%%%%%%%%%%%%%%%%%%%%%%%%%%%%%%%%

However, this is not sufficient to compute the curvature. To express curvature explicitly in terms of the metric, two additional components are required: the first derivative of the metric tensor—since the Christoffel symbols are defined using both the metric and its derivatives—and the second derivative of the metric tensor, as curvature depends on the derivatives of the Christoffel symbols. Therefore, we need to identify two additional fundamental building blocks to compute the curvature tensor. The first derivative of the metric tensor can be expressed as a combination of the Jacobian and the derivative of the Jacobian. Thus, the next step is to derive the explicit form of the Jacobian's derivative.
\begin{equation}
\begin{array}{ll}
    \frac{\partial^2 x^{(n+1)i}}{\partial x^{(n)k} \partial x^{(n)j}} =& W^{(n+1)i}_{\phantom{(n+1)i}j}W^{(n+1)a_2}_{\phantom{(n+1)a_2}k}E^{i}_{\phantom{i}a_2}(\sigma^2)^{i}+ W^{(n+1)i}_{\phantom{(n+1)i}k}W^{(n+1)a_3}_{\phantom{(n+1)a_3}j}E^{i}_{\phantom{i}a_3}(\sigma^2)^{i}\\
    & - (W^{(n+1)}x^{(n)} + b^{(n+1)})^{i}W^{(n+1)a_3}_{\phantom{(n+1)a_3}j}W^{(n+1)i}_{\phantom{(n+1)i}k}E^{i}_{\phantom{i}a_3}(\sigma^2)^{i}\\
    & + 2(W^{(n+1)}x^{(n)} + b^{(n+1)})^{i}W^{(n+1)a_3}_{\phantom{(n+1)a_3}j}E^{i}_{\phantom{i}a_3}W^{(n+1)a_6}_{\phantom{(n+1)a_6}k}E^{i}_{\phantom{i}a_6}(\sigma^3)^{i}
\end{array}
\end{equation}
Finally, the derivative of the Christoffel symbols can be induced with the second derivative of the Jacobian.
\begin{equation}
\begin{array}{l}
    \frac{\partial^3 x^{(n+1)i}}{\partial x^{(n)l} \partial x^{(n)k}\partial x^{(n)j}} = \\
    \qquad
       -2W^{(n+1)i}_{\phantom{(n+1)i}l}W^{(n+1)a_2}_{\phantom{(n+1)a_2}k}W^{(n+1)i}_{\phantom{(n+1)i}j}E^{i}_{\phantom{i}a_2}(\sigma^2)^{i} -W^{(n+1)i}_{\phantom{(n+1)i}j}W^{(n+1)a_3}_{\phantom{(n+1)a_3}l}W^{(n+1)i}_{\phantom{(n+1)i}k}E^{i}_{\phantom{i}a_3}(\sigma^2)^{i} \\
      \qquad + 4W^{(n+1)i}_{\phantom{(n+1)i}l}W^{(n+1)a_2}_{\phantom{(n+1)a_2}k}W^{(n+1)a_9}_{\phantom{(n+1)a_9}j}E^{i}_{\phantom{i}a_2}E^{i}_{\phantom{i}a_9}(\sigma^3)^{i}\\
       \qquad+ (W^{(n+1)}x^{(n)} + b^{(n+1)})^{i}W^{(n+1)a_3}_{\phantom{(n+1)a_3}l}W^{(n+1)i}_{\phantom{(n+1)i}k}W^{(n+1)i}_{\phantom{(n+1)i}j}E^{i}_{\phantom{i}a_3}(\sigma^2)^{i}\\
       \qquad- 2(W^{(n+1)}x^{(n)} + b^{(n+1)})^{i}W^{(n+1)a_3}_{\phantom{(n+1)a_3}l}W^{(n+1)i}_{\phantom{(n+1)i}k}W^{(n+1)a_9}_{\phantom{(n+1)a_9}j}E^{i}_{\phantom{i}a_3}E^{i}_{\phantom{i}a_9}(\sigma^3)^{i}\\
       \qquad+ 2W^{(n+1)i}_{\phantom{(n+1)i}j}W^{(n+1)a_3}_{\phantom{(n+1)a_3}l}W^{(n+1)a_6}_{\phantom{(n+1)a_6}k}E^{i}_{\phantom{i}a_3}E^{i}_{\phantom{i}a_6}(\sigma^3)^{i}\\
       % & - 2(W^{(n+1)}x + b^{(n+1)})^{i}W^{(n+1)a_3}_{\phantom{(n+1)a_3}m}W^{(n+1)a_9}_{\phantom{(n+1)a_9}j}W^{(n+1)a_6}_{\phantom{(n+1)a_6}k} \delta^{i}_{\phantom{i}a_9 a_{10}} E^{a_{10}}_{\phantom{a_{10}}a_3}  E^{a_{1}}_{\phantom{a_{1}}a_6} (\sigma^3)^{i}\\
       \qquad- 4(W^{(n+1)}x^{(n)} + b^{(n+1)})^{i}W^{(n+1)a_3}_{\phantom{(n+1)a_3}l} W^{(n+1)a_6}_{\phantom{(n+1)a_6}k} W^{(n+1)i}_{\phantom{(n+1)i}j} E^{i}_{\phantom{i}(a_3} E^{i}_{\phantom{i}a_6)} (\sigma^3)^{i}\\
       \qquad+ 6(W^{(n+1)}x^{(n)} + b^{(n+1)})^{i}W^{(n+1)a_3}_{\phantom{(n+1)a_3}l}W^{(n+1)a_6}_{\phantom{(n+1)a_6}k}W^{(n+1)a_9}_{\phantom{(n+1)a_9}j}E^{i}_{\phantom{i}a_9}E^{i}_{\phantom{i}a_3}E^{i}_{\phantom{i}a_6}(\sigma^4)^{i}
\end{array}
\end{equation}
By gathering and utilizing the three building blocks described above and imposing them into Eq.~\ref{indmet}, \ref{affine_connection}, \ref{Riemann_curv} and \ref{scalar_curv} , the scalar curvature of the given curved space can be explicitly calculated.

\subsection{Model architecture}
Our model is designed to perform effectively in a two-task setting, regardless of whether the data distributions between tasks are balanced or unbalanced. Therefore, the basic architecture consists of two distinct task-specific models connected by a transfer module, as shown in Figure~\ref{fig:algo_figure}. These task-specific models are connected only through the curvature matching section; thus, their architectures are fully flexible, with the sole constraint that the dimensions of the latent vectors fed into the transfer module must match. This allows each task-specific model to be independently designed, taking into account the complexity of the task and its corresponding data distribution.

When an input data point is fed into the model, the first step is to construct an embedding vector from the input information. We denote the embedding vector as $z_t$ for the target task and $z_s$ for the source task. These embedding vectors follow two distinct paths in the architecture: one path leads to the transfer module, and the other proceeds to the head module in the model. The transfer module maps each embedding to a vector of the same dimension in a locally flat coordinate frame. 
\begin{gather}
    z' = \mathrm{Transfer}(z), \qquad \hat{z} = \mathrm{Transfer}^{-1}(z')
\end{gather}
However, for the inverse transfer, direct computation of the inverse matrix during backpropagation can be unstable. To address this, we designed an autoencoder architecture to map the embedding vector from the locally flat frame back to the original space. Accordingly, we define an autoencoder loss to guide this reconstruction process.
\begin{equation}
    l_{\mathrm{auto}} = \mathrm{MSE}(z, \hat{z})
\end{equation}
Since the transferred vectors $z'_s$ and $z'_t$ originate from the same input, they should match—assuming the coordinate systems are aligned, i.e., expressed in a common locally flat frame.
\begin{gather}
    z'_s = \mathrm{Transfer}_{s\rightarrow LF}(z_s) \qquad    \hat{z}_s = \mathrm{Transfer}^{-1}_{\phantom{-1}LF \rightarrow s}(z'_s)\\
    z'_t = \mathrm{Transfer}_{t\rightarrow LF}(z_t) \qquad    \hat{z}_t = \mathrm{Transfer}^{-1}_{\phantom{-1}LF \rightarrow t}(z'_t)
\end{gather}
Here, $\mathrm{Model}_{s \rightarrow LF}$ denotes the transfer module that maps the embedding vector from the source space to the locally flat (LF) frame, and vice versa. To encourage alignment, we introduce a consistency loss by matching the embedding vectors from both the source and target task models within the shared locally flat frame.
\begin{equation}
    l_{cons} = \mathrm{MSE}(z'_s, z'_t)
\end{equation}
To further reinforce the connection between the source and target tasks, we introduce an additional loss—the mapping loss—which aligns the downstream prediction of the original target label with the prediction obtained from an embedding vector transferred from the source model.
\begin{equation}
    l_{map} = \mathrm{MSE}(y_t, \hat{y}_{s \rightarrow t})
\end{equation}
Here, $y_t$ denotes the target label and $\hat{y}_{s \rightarrow t}$ represents the predicted value obtained from the embedding vector transferred from the source model. And the ordinary regression loss is also important.
\begin{equation}
    l_{reg} = \mathrm{MSE}(y_t, \hat{y}_t)
\end{equation}

Finally, we define the curvature and metric losses. The metric loss plays a crucial role, as the space formed by the transfer module lacks any form of direct supervision. Without proper regularization, the space is not guaranteed to be locally flat, since there are infinitely many ways to define a basis that still satisfy the previously introduced constraints. The metric loss guides the transfer module toward preserving local flatness. It is defined as the discrepancy between the induced flat metric and the Euclidean metric, which in this case is represented by the identity matrix $\eta_{ij}$.
\begin{eqnarray}
    l_{metric} = \mathrm{MSE}(\eta_{ij}, {\eta_{(s)}}_{ij}) + \mathrm{MSE}(\eta_{ij}, {\eta_{(t)}}_{ij})\\
    \eta_{(s)ij} = \left(\frac{\partial {\hat{z}}{_s}}{\partial {\hat{z}'}_s}\right)^{m}_{i}\left(\frac{\partial {z'}_s}{\partial {\hat{z}}_{s}}\right)^{k}_{m} \eta_{kl} \left(\frac{\partial {z'}_s}{\partial {\hat{z}}_{s}}\right)^{l}_{n}\left(\frac{\partial {\hat{z}}{_s}}{\partial {\hat{z}'}_s}\right)^{n}_{j}\\
    \eta_{(t)ij} = \left(\frac{\partial {\hat{z}}{_t}}{\partial {\hat{z}'}_t}\right)^{m}_{i}\left(\frac{\partial {z'}_t}{\partial {\hat{z}}_{t}}\right)^{k}_{m} \eta_{kl} \left(\frac{\partial {z'}_t}{\partial {\hat{z}}_{t}}\right)^{l}_{n}\left(\frac{\partial {\hat{z}}{_t}}{\partial {\hat{z}'}_t}\right)^{n}_{j}
\end{eqnarray}
${\eta_{(s)}}_{ij}$ and ${\eta_{(t)}}_{ij}$ are the induced flat metrics obtained by loop computations with the inverse transfer from the source and target, respectively, to the transfer module. The symbol $\hat{z}'$ denotes the transferred latent vector, which is mapped from the flat space to the curved space, and then back to the flat space.

The final loss term is the \textit{curvature matching loss}. Since we have already introduced the fundamental building blocks for computing scalar curvature in Sections~\ref{preliminary} and~\ref{strategy}, the scalar curvature can now be computed analytically. Given that the curvatures of the target and source spaces should align, we define this curvature matching loss as the most critical and final component of our architecture.
\begin{equation}
    l_{curv} = \mathrm{MSE}(R_s, R_t)
\end{equation}
Where $R_t$ and $R_s$ are the Ricci scalar curvatures from the target and the source respectively. Finally, by combining all with appropriate hyperparameters, the main loss of the model can be defined.
\begin{equation}
    l_{tot} = l_{reg} + \alpha l_{auto} + \beta l_{cons} + \gamma l_{map} + \delta l_{metric}  + \epsilon l_{curv}
\end{equation}
Each hyperparameter can be tuned individually to enhance the model's predictive performance. In particular, the weight for the metric loss should be increased, as its magnitude is significantly smaller compared to the other loss terms. The specific configurations of these loss components, along with the model parameters, are detailed in Appendix~\ref{architecture}. Furthermore, detailed schematics of our model are provided in Appendix Figure~\ref{fig:apdx_algo}.

%%%%%%%%%%%%%%%%%%%%%%%%%%%%%%%%%%%%%%%%%%%%%%%%%%%%%%%%%%%%%%%%%%%%%%%%%%
 %%%%%%%%%%%%%%%%%%%%%%%%%%%%%%%%%%%%%%%%%%%%%%%%%%%%%%%%%%%%%%%%%%%%%%%%%%
 %%%%%%%%%%%%%%%%%%%%%%%%%%%%%%%%%%%%%%%%%%%%%%%%%%%%%%%%%%%%%%%%%%%%%%%%%%

\section{Experiments}
\label{experiments}

% In this section, we present the detailed experimental setups, main results, and several ablation studies. 

\subsection{Experimental settings}
The experiments are conducted using open datasets from three distinct databases (OCHEM \citet{sushko2011online}, PubChem \citet{10.1093/nar/gkac956}, and CCCB \citet{cccb}) forming 23 task pairs across 14 different tasks using two distinct data splitting schemes: the conventional random split and the scaffold-based split, the latter of which simulates OOD scenarios. A detailed explanation of these datasets and their corresponding prediction tasks is provided in the Appendix~\ref{detailed_setup}. To ensure the robustness of the results, all experiments are performed using 4-fold cross-validation. Each experiment is run on a single NVIDIA A40 GPU.

To evaluate our method, we compare it against several benchmark models, including single-task learning (STL), MTL, global structure preserving loss-based knowledge distillation (GSP-KD) \citet{joshi2022representation}, two variants of TL (retraining all layers vs. retraining the head only), and GATE \citet{ko2023geometricallyalignedtransferencoder}. We ensure fairness by maintaining the same encoder and head architectures across all benchmark models and our method. Detailed hyperparameter configurations are in the Appendix~\ref{architecture}.

\begin{figure}[ht]
\begin{center}
\includegraphics[clip, width=1\linewidth]{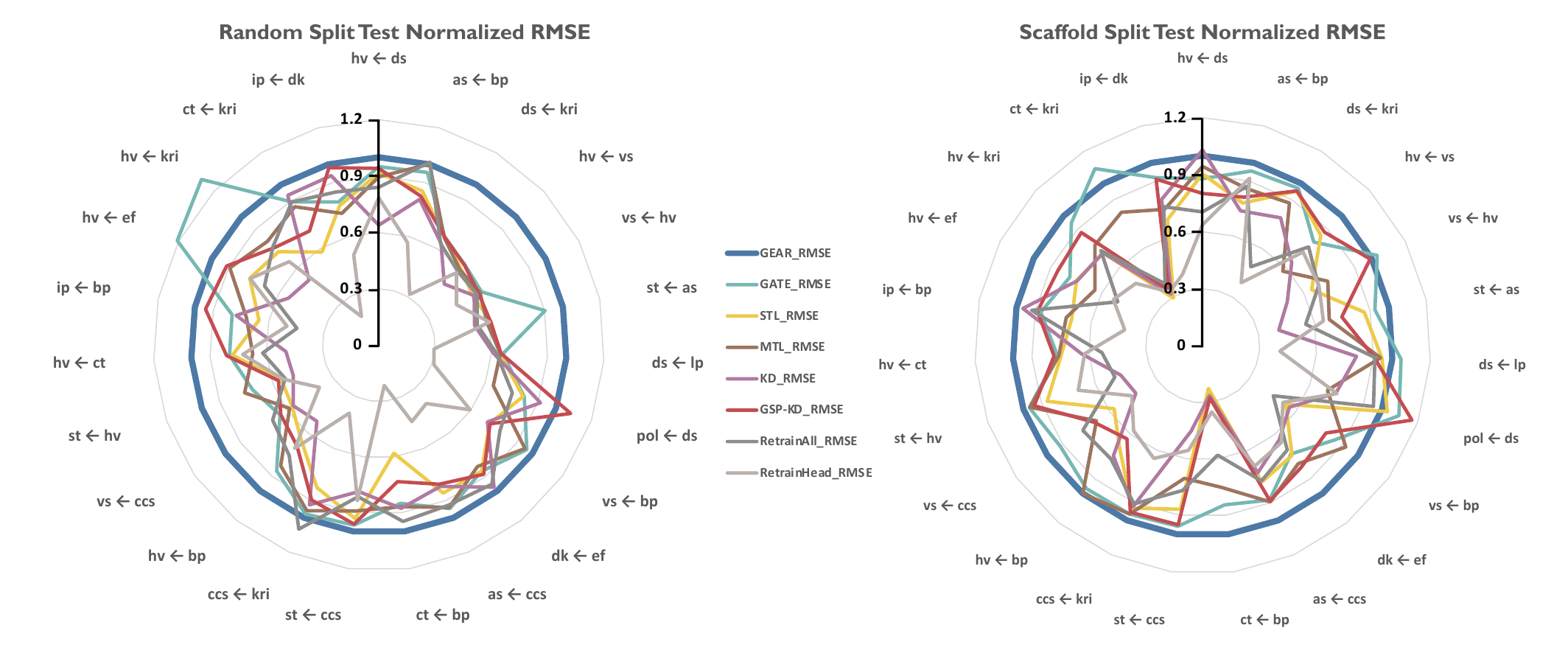}
\end{center}
\caption{The results are illustrated in the form of a radar chart. The baseline in the chart corresponds to the RMSE performance of GEAR, which is normalized to 1 (higher is better). The performances of other benchmark models are represented as ratios relative to the GEAR RMSE values. Due to space constraints, the detailed experimental results are provided in Table~[\ref{random_result_1}, \ref{random_result_2}, \ref{scaffold_part1}, \ref{scaffold_part2}] in the Appendix~\ref{bare_result}}
\label{fig:main_figure}
\end{figure}

\subsection{Main results}
Figure~\ref{fig:main_figure} demonstrates the superior performance of our algorithm compared to other benchmark models. In both data split schemes, our model consistently outperforms the baseline models by considerable margins. Notably, when counting the number of best-performing tasks, GEAR achieves the lowest RMSE in 18 out of 23 task pairs under the random split and in 17 out of 23 under the scaffold split. Furthermore, when including second-best performances, GEAR ranks within the top two in 22 out of 23 for both split schemes.

From a performance standpoint, GEAR improves the average RMSE over GATE by $14.4 \%$ in the random split and by $8.3 \%$ in the scaffold split. Compared to the third-best model, GEAR achieves an improvement of $22.8 \%$ (MTL) under the random split and $21.4 \%$ (GSP-KD) under the scaffold split.

%%%%%%%%%%%%%%%%%%%%%%%%%%%%%%%%%%%%%%%%%%%%%%%%%%%%%%%%%%%%%%%%%%%%%%%%%%
 %%%%%%%%%%%%%%%%%%%%%%%%%%%%%%%%%%%%%%%%%%%%%%%%%%%%%%%%%%%%%%%%%%%%%%%%%%
 %%%%%%%%%%%%%%%%%%%%%%%%%%%%%%%%%%%%%%%%%%%%%%%%%%%%%%%%%%%%%%%%%%%%%%%%%%

\section{Ablation studies}
% In this section, we present several ablation studies focusing on the newly introduced curvature loss term. First, we describe the role and contribution of the curvature loss. Then, we demonstrate the robustness of GEAR under out-of-distribution (OOD) datasets. Finally, we analyze the computational cost associated with curvature computation to explain why autograd-based approaches were not employed in this work.
\subsection{Role of curvature loss}
Since GEAR is constructed under a TL scheme, it is crucial to verify that the loss terms connecting the source and target tasks effectively facilitate information transfer. To support this claim, we conducted three different experiments and plotted training and validation accuracy curves.

\begin{figure}[ht]
\begin{center}
\includegraphics[trim={1cm 7.5cm 1cm 7.5cm}, clip, width=1\linewidth]{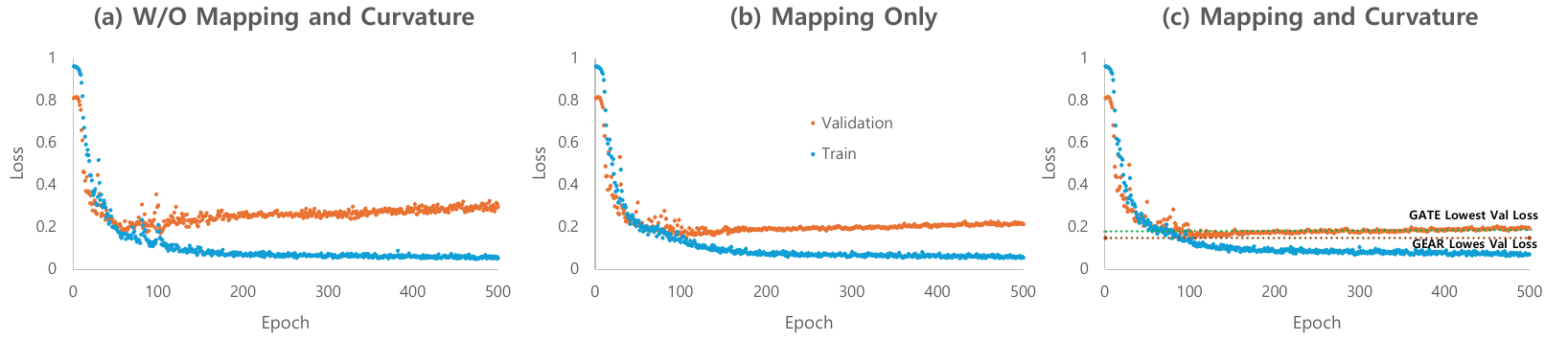}
\end{center}
\caption{These plots illustrate the primary role of the curvature loss. In figure (a), both the mapping loss and the curvature loss are turned off. In figure (b), only the mapping loss is enabled. In figure (c), both the mapping and curvature losses are activated.}
\label{fig:loss_figure}
\end{figure}

As shown in Figure~\ref{fig:loss_figure}, when both the mapping and curvature matching losses are turned off, the loss curve exhibits a severe overfitting issue. Enabling the mapping loss alone helps to stabilize this overfitting to some extent. However, when both losses are activated, overfitting is significantly suppressed, and the validation loss reaches the lowest value overall.

For comparison, we included the minimum validation loss value of GATE as a green dotted line, alongside that of GEAR (in brown dotted line) under the same experimental setting. The comparison reveals that GEAR achieves a lower minimum validation loss than GATE, with a performance improvement margin of $17.5 \%$.

\begin{figure}[ht]
\begin{center}
\includegraphics[trim={0cm 7cm 0cm 7cm}, clip, width=1\linewidth]{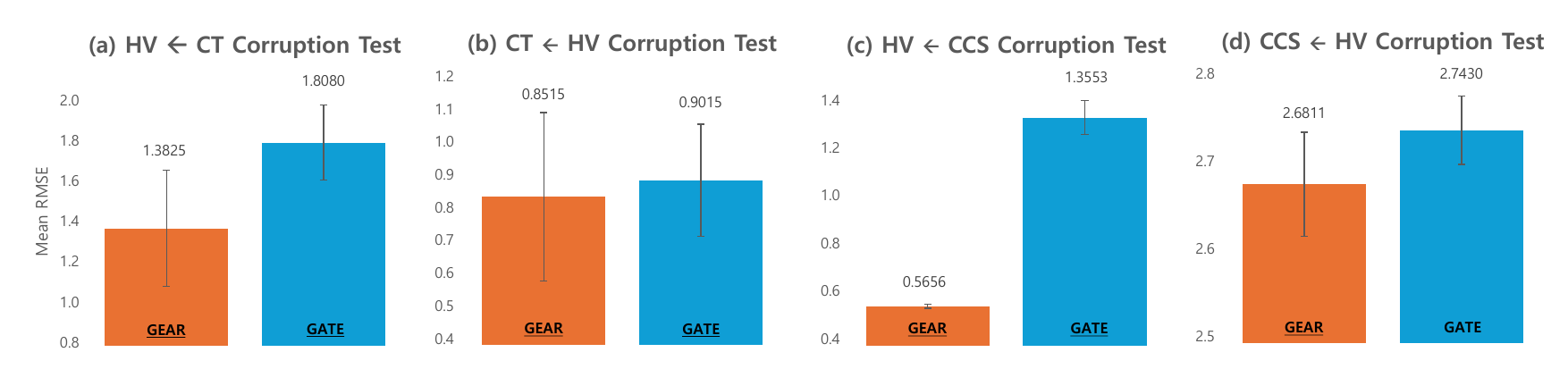}
\end{center}
\caption{This figure highlights the superior performance of GEAR in noisy data prediction tasks. The reported values represent the average RMSE across all four folds, with error bars indicating the standard deviation. Specifically, Figure (a) shows HV prediction results using CT as the source, (b) shows CT prediction results using HV as the source, (c) shows HV prediction results using CCS as the source, and (d) shows CCS prediction results using HV as the source.}
\label{fig:corruption_figure}
\end{figure}

\subsection{Robustness under noisy dataset}
In this subsection, we demonstrate the robustness of GEAR under noisy conditions on dataset to evaluate the model’s regularization effect. To support this claim, we constructed a modified dataset by corrupting data points with values at least twice the standard deviation of each dataset. Specifically, we selected $10 \%$ of the test set containing values greater than the dataset’s standard deviation. These selected labels were corrupted by multiplying them by -1 and then injected into the training dataset. After training, we evaluated the model by feeding these corrupted samples and comparing the predictions against their original, uncorrupted labels. This procedure was repeated under a 4-fold cross-validation scheme to ensure the reliability of the results.

As shown in Figure~\ref{fig:corruption_figure}, GEAR consistently outperforms conventional models across all cases.

\subsection{Computational costs}
\begin{figure}[ht]
\centering
\includegraphics[trim={0cm 0.3cm 0cm 0.2cm}, clip, width=0.8\linewidth]{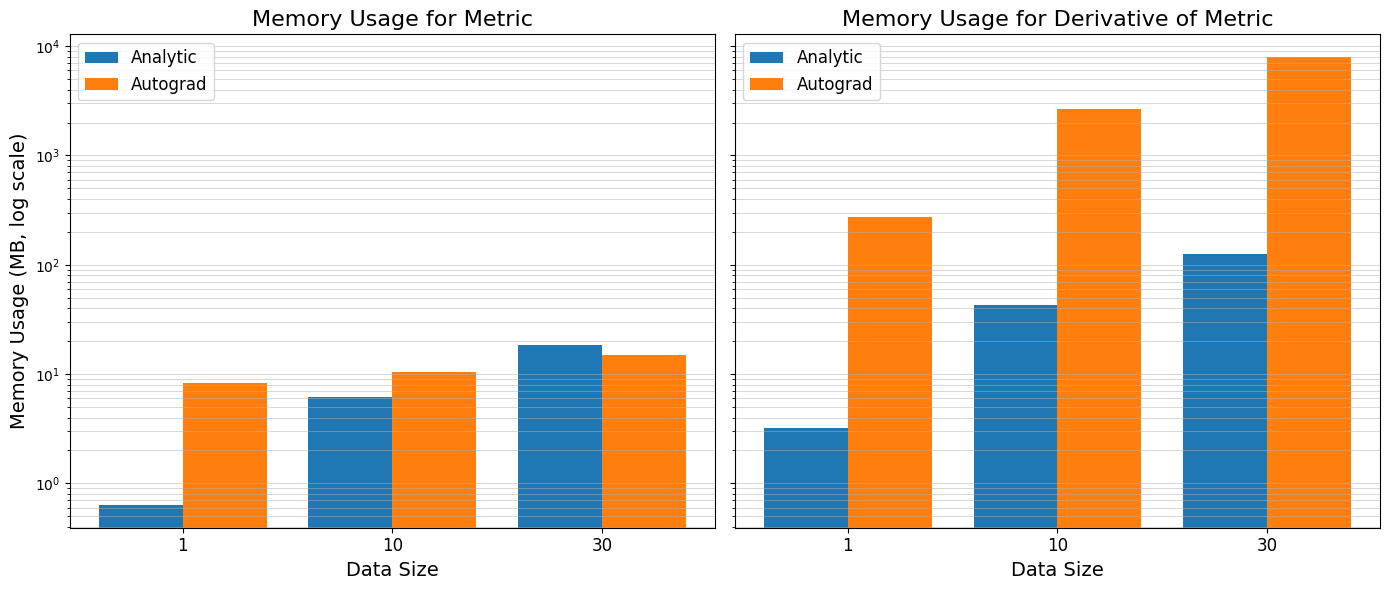}
\caption{Memory usage was visualized in the form of bar charts in log scale. The charts compare the memory consumption of the analytic and autograd-based methods when computing the metric and the derivative of the metric, evaluated at data sizes of 1, 10, and 30. The corresponding bars are labeled as Analytic and Autograd, respectively.}
\label{fig:fig3}
\end{figure}
In this subsection, we demonstrate the necessity of computing the curvature analytically.
% Although this approach is mathematically complex, it can be more easily implemented using PyTorch’s autograd. 
Although metric and its derivatives can be calculated by PyTorch's autograd, autograd requires substantial computational cost compared to analytic computation. We compared memory consumption between the analytical and autograd-based methods when computing the metric and its derivatives across varying data sizes.

As shown in Figure \ref{fig:fig3}, both methods exhibited similar memory usage for metric computation, which involves first-order derivatives. However, for computing metric derivatives (i.e., second-order), the autograd approach consumed approximately 85.5× more memory at data size 1. Due to this overhead, autograd-based training was infeasible under our GPU constraints. In contrast, the analytical method enabled fast and memory-efficient training, requiring only 0.5 seconds per iteration at batch size 512.

%%%%%%%%%%%%%%%%%%%%%%%%%%%%%%%%%%%%%%%%%%%%%%%%%%%%%%%%%%%%%%%%%%%%%%%%%%
 %%%%%%%%%%%%%%%%%%%%%%%%%%%%%%%%%%%%%%%%%%%%%%%%%%%%%%%%%%%%%%%%%%%%%%%%%%
 %%%%%%%%%%%%%%%%%%%%%%%%%%%%%%%%%%%%%%%%%%%%%%%%%%%%%%%%%%%%%%%%%%%%%%%%%%

\section{Discussion}
We introduced a novel TL algorithm, GEAR, based on Riemannian differential geometry. Since deep learning models are inherently smooth and differentiable, the Jacobian of the transfer module can be computed analytically. From the Jacobian, the induced curved metric can be derived and used for curvature computation. The Ricci scalar curvature encapsulates the full geometric characteristics of the latent space; by matching the curvature between the target and source tasks, the latent spaces can be accurately aligned. Experimental results on 23 pairs of molecular property prediction tasks demonstrate the superior performance of GEAR compared to benchmark models.

Simplifying or retracting curvature matching—without compromising generality—can help alleviate the implementation complexity and computational overhead associated with curvature computation. GEAR further contributes structural flexibility by linking source and target tasks via transfer modules, without constraining the downstream architecture. This design keeps encoder modules fully unconstrained, facilitating smooth adaptation to multi-modal learning scenarios. Additionally, the framework is inherently extensible to settings involving more than two interrelated tasks, paving the way for broader applications in multi-task transfer learning.

%%%%%%%%%%%%%%%%%%%%%%%%%%%%%%%%%%%%%%%%%%%%%%%%%%%%%%%%%%%%

\bibliographystyle{plainnat}
\bibliography{neurips_2025}

\begin{thebibliography}{44}
\providecommand{\natexlab}[1]{#1}
\providecommand{\url}[1]{\texttt{#1}}
\expandafter\ifx\csname urlstyle\endcsname\relax
  \providecommand{\doi}[1]{doi: #1}\else
  \providecommand{\doi}{doi: \begingroup \urlstyle{rm}\Url}\fi

\bibitem[Basu et~al.(2023)Basu, Katdare, Sattigeri, Chenthamarakshan, Driggs-Campbell, Das, and Varshney]{basu2023efficientequivarianttransferlearning}
Sourya Basu, Pulkit Katdare, Prasanna Sattigeri, Vijil Chenthamarakshan, Katherine Driggs-Campbell, Payel Das, and Lav~R. Varshney.
\newblock Efficient equivariant transfer learning from pretrained models, 2023.
\newblock URL \url{https://arxiv.org/abs/2305.09900}.

\bibitem[Bronstein et~al.(2017)Bronstein, Bruna, LeCun, Szlam, and Vandergheynst]{bronstein2017geometric}
Michael~M Bronstein, Joan Bruna, Yann LeCun, Arthur Szlam, and Pierre Vandergheynst.
\newblock Geometric deep learning: going beyond euclidean data.
\newblock \emph{IEEE Signal Processing Magazine}, 34\penalty0 (4):\penalty0 18--42, 2017.

\bibitem[Bruna et~al.(2013)Bruna, Zaremba, Szlam, and Lecun]{bruna2013}
Joan Bruna, Wojciech Zaremba, Arthur Szlam, and Yann Lecun.
\newblock Spectral networks and locally connected networks on graphs.
\newblock 12 2013.

\bibitem[Buterez et~al.(2024)Buterez, Janet, Kiddle, Oglic, and Li{\'o}]{buterez2024transfer}
David Buterez, Jon~Paul Janet, Steven~J Kiddle, Dino Oglic, and Pietro Li{\'o}.
\newblock Transfer learning with graph neural networks for improved molecular property prediction in the multi-fidelity setting.
\newblock \emph{Nature communications}, 15\penalty0 (1):\penalty0 1517, 2024.

\bibitem[Coley et~al.(2019)Coley, Jin, Rogers, Jamison, Jaakkola, Green, Barzilay, and Jensen]{C8SC04228D}
Connor W. Coley, Wengong Jin, Luke Rogers, Timothy~F. Jamison, Tommi~S. Jaakkola, William~H. Green, Regina Barzilay, and Klavs~F. Jensen.
\newblock A graph-convolutional neural network model for the prediction of chemical reactivity.
\newblock \emph{Chem. Sci.}, 10:\penalty0 370--377, 2019.
\newblock \doi{10.1039/C8SC04228D}.
\newblock URL \url{http://dx.doi.org/10.1039/C8SC04228D}.

\bibitem[Defferrard et~al.(2016)Defferrard, Bresson, and Vandergheynst]{defferrard2016}
Michaël Defferrard, Xavier Bresson, and Pierre Vandergheynst.
\newblock Convolutional neural networks on graphs with fast localized spectral filtering.
\newblock 06 2016.

\bibitem[Duvenaud et~al.(2015)Duvenaud, Maclaurin, Aguilera-Iparraguirre, Gómez-Bombarelli, Hirzel, Aspuru-Guzik, and Adams]{duvenaud2015}
David Duvenaud, Dougal Maclaurin, Jorge Aguilera-Iparraguirre, Rafael Gómez-Bombarelli, Timothy Hirzel, Alán Aspuru-Guzik, and Ryan Adams.
\newblock Convolutional networks on graphs for learning molecular fingerprints.
\newblock \emph{Advances in Neural Information Processing Systems (NIPS)}, 13, 09 2015.

\bibitem[Falk et~al.(2023)Falk, Bonati, Novelli, Parrinello, and Pontil]{falk2023transfer}
John Falk, Luigi Bonati, Pietro Novelli, Michele Parrinello, and Massimiliano Pontil.
\newblock Transfer learning for atomistic simulations using gnns and kernel mean embeddings.
\newblock \emph{Advances in Neural Information Processing Systems}, 36:\penalty0 29783--29797, 2023.

\bibitem[Grattarola et~al.(2019)Grattarola, Livi, and Alippi]{grattarola2019adversarial}
Daniele Grattarola, Lorenzo Livi, and Cesare Alippi.
\newblock Adversarial autoencoders with constant-curvature latent manifolds.
\newblock \emph{Applied Soft Computing}, 81:\penalty0 105511, 2019.

\bibitem[Hoffmann et~al.(2023)Hoffmann, Schmidt, Botti, and Marques]{hoffmann2023transfer}
Noah Hoffmann, Jonathan Schmidt, Silvana Botti, and Miguel~AL Marques.
\newblock Transfer learning on large datasets for the accurate prediction of material properties.
\newblock \emph{Digital Discovery}, 2\penalty0 (5):\penalty0 1368--1379, 2023.

\bibitem[Hu et~al.(2024)Hu, Zhan, Tian, Xiong, and Lu]{hu2024enhanced}
Wenbo Hu, Hongjian Zhan, Yinghong Tian, Yujie Xiong, and Yue Lu.
\newblock Enhanced video clustering using multiple riemannian manifold-valued descriptors and audio-visual information.
\newblock \emph{Expert Systems with Applications}, 246:\penalty0 123099, 2024.

\bibitem[III(2022)]{cccb}
Russell D.~Johnson III.
\newblock Nist computational chemistry comparison and benchmark database.
\newblock \emph{NIST Standard Reference Database}, 101, 2022.
\newblock URL \url{http://cccbdb.nist.gov/}.

\bibitem[Jin et~al.(2018)Jin, Yang, Barzilay, and Jaakkola]{jin2018}
Wengong Jin, Kevin Yang, Regina Barzilay, and Tommi Jaakkola.
\newblock Learning multimodal graph-to-graph translation for molecular optimization, 12 2018.

\bibitem[Joshi et~al.(2022)Joshi, Liu, Xun, Lin, and Foo]{joshi2022representation}
Chaitanya~K Joshi, Fayao Liu, Xu~Xun, Jie Lin, and Chuan~Sheng Foo.
\newblock On representation knowledge distillation for graph neural networks.
\newblock \emph{IEEE Transactions on Neural Networks and Learning Systems}, 2022.

\bibitem[Kim et~al.(2022)Kim, Chen, Cheng, Gindulyte, He, He, Li, Shoemaker, Thiessen, Yu, Zaslavsky, Zhang, and Bolton]{10.1093/nar/gkac956}
Sunghwan Kim, Jie Chen, Tiejun Cheng, Asta Gindulyte, Jia He, Siqian He, Qingliang Li, Benjamin~A Shoemaker, Paul~A Thiessen, Bo~Yu, Leonid Zaslavsky, Jian Zhang, and Evan~E Bolton.
\newblock {PubChem 2023 update}.
\newblock \emph{Nucleic Acids Research}, 51\penalty0 (D1):\penalty0 D1373--D1380, 10 2022.
\newblock ISSN 0305-1048.
\newblock \doi{10.1093/nar/gkac956}.
\newblock URL \url{https://doi.org/10.1093/nar/gkac956}.

\bibitem[Ko et~al.(2023{\natexlab{a}})Ko, Cho, Jeong, Han, Lee, and Lee]{ko2023grouping}
Sung~Moon Ko, Sungjun Cho, Dae-Woong Jeong, Sehui Han, Moontae Lee, and Honglak Lee.
\newblock Grouping matrix based graph pooling with adaptive number of clusters.
\newblock \emph{Proceedings of the AAAI Conference on Artificial Intelligence}, 37\penalty0 (7):\penalty0 8334–8342, June 2023{\natexlab{a}}.
\newblock ISSN 2159-5399.
\newblock \doi{10.1609/aaai.v37i7.26005}.
\newblock URL \url{http://dx.doi.org/10.1609/aaai.v37i7.26005}.

\bibitem[Ko et~al.(2023{\natexlab{b}})Ko, Lee, Jeong, Lim, and Han]{ko2023geometricallyalignedtransferencoder}
Sung~Moon Ko, Sumin Lee, Dae-Woong Jeong, Woohyung Lim, and Sehui Han.
\newblock Geometrically aligned transfer encoder for inductive transfer in regression tasks, 2023{\natexlab{b}}.
\newblock URL \url{https://arxiv.org/abs/2310.06369}.

\bibitem[Ko et~al.(2024)Ko, Lee, Jeong, Kim, Lee, Yim, and Han]{ko2024multitaskextensiongeometricallyaligned}
Sung~Moon Ko, Sumin Lee, Dae-Woong Jeong, Hyunseung Kim, Chanhui Lee, Soorin Yim, and Sehui Han.
\newblock Multitask extension of geometrically aligned transfer encoder, 2024.
\newblock URL \url{https://arxiv.org/abs/2405.01974}.

\bibitem[Kulis et~al.(2011)Kulis, Saenko, and Darrell]{Kulis2011WhatYS}
Brian Kulis, Kate Saenko, and Trevor Darrell.
\newblock What you saw is not what you get: Domain adaptation using asymmetric kernel transforms.
\newblock \emph{CVPR 2011}, pages 1785--1792, 2011.
\newblock URL \url{https://api.semanticscholar.org/CorpusID:7419723}.

\bibitem[Lee et~al.(2024)Lee, Jeong, Ko, Lee, Kim, Yim, Han, Kim, and Lim]{lee2024scalablemultitasktransferlearning}
Chanhui Lee, Dae-Woong Jeong, Sung~Moon Ko, Sumin Lee, Hyunseung Kim, Soorin Yim, Sehui Han, Sungwoong Kim, and Sungbin Lim.
\newblock Scalable multi-task transfer learning for molecular property prediction, 2024.
\newblock URL \url{https://arxiv.org/abs/2410.00432}.

\bibitem[Lee et~al.(2022)Lee, Kim, Choi, and Park]{lee2022statistical}
Yonghyeon Lee, Seungyeon Kim, Jinwon Choi, and Frank Park.
\newblock A statistical manifold framework for point cloud data.
\newblock In \emph{International Conference on Machine Learning}, pages 12378--12402. PMLR, 2022.

\bibitem[Li et~al.(2023)Li, Fei, Wang, Shan, and Lu]{li2023geometry}
Yangyang Li, Chaoqun Fei, Chuanqing Wang, Hongming Shan, and Ruqian Lu.
\newblock Geometry flow-based deep riemannian metric learning.
\newblock \emph{IEEE/CAA Journal of Automatica Sinica}, 10\penalty0 (9):\penalty0 1882--1892, 2023.

\bibitem[Long et~al.()Long, Wang, Ding, Cheng, Zhang, and Wang]{doi:10.1137/1.9781611972825.47}
Mingsheng Long, Jianmin Wang, Guiguang Ding, Wei Cheng, Xiang Zhang, and Wei Wang.
\newblock \emph{Dual Transfer Learning}, pages 540--551.
\newblock \doi{10.1137/1.9781611972825.47}.
\newblock URL \url{https://epubs.siam.org/doi/abs/10.1137/1.9781611972825.47}.

\bibitem[Loshchilov and Hutter(2017)]{loshchilov2017decoupled}
Ilya Loshchilov and Frank Hutter.
\newblock Decoupled weight decay regularization.
\newblock \emph{arXiv preprint arXiv:1711.05101}, 2017.

\bibitem[Pan et~al.(2020)Pan, Cui, Duy~Le, Li, and Zhang]{9051683}
Jianhan Pan, Teng Cui, Thuc Duy~Le, Xiaomei Li, and Jing Zhang.
\newblock Multi-group transfer learning on multiple latent spaces for text classification.
\newblock \emph{IEEE Access}, 8:\penalty0 64120--64130, 2020.
\newblock \doi{10.1109/ACCESS.2020.2984571}.

\bibitem[Park et~al.(2023)Park, Kwon, Choi, Jo, and Uh]{park2023understanding}
Yong-Hyun Park, Mingi Kwon, Jaewoong Choi, Junghyo Jo, and Youngjung Uh.
\newblock Understanding the latent space of diffusion models through the lens of riemannian geometry.
\newblock \emph{Advances in Neural Information Processing Systems}, 36:\penalty0 24129--24142, 2023.

\bibitem[Pegios et~al.(2024)Pegios, Feragen, Hansen, and Arvanitidis]{pegios2024counterfactual}
Paraskevas Pegios, Aasa Feragen, Andreas~Abildtrup Hansen, and Georgios Arvanitidis.
\newblock Counterfactual explanations via riemannian latent space traversal.
\newblock \emph{CoRR}, 2024.

\bibitem[Peng et~al.(2021)Peng, Li, Wamsley, Wei, and Roeder]{doi:10.1073/pnas.2024383118}
Minshi Peng, Yue Li, Brie Wamsley, Yuting Wei, and Kathryn Roeder.
\newblock Integration and transfer learning of single-cell transcriptomes via cfit.
\newblock \emph{Proceedings of the National Academy of Sciences}, 118\penalty0 (10):\penalty0 e2024383118, 2021.
\newblock \doi{10.1073/pnas.2024383118}.
\newblock URL \url{https://www.pnas.org/doi/abs/10.1073/pnas.2024383118}.

\bibitem[Quattoni et~al.(2008)Quattoni, Collins, and Darrell]{Quattoni}
Ariadna Quattoni, Michael Collins, and Trevor Darrell.
\newblock Transfer learning for image classification with sparse prototype representations.
\newblock \emph{Proceedings / CVPR, IEEE Computer Society Conference on Computer Vision and Pattern Recognition. IEEE Computer Society Conference on Computer Vision and Pattern Recognition}, 2, 03 2008.
\newblock \doi{10.1109/CVPR.2008.4587637}.

\bibitem[Radhakrishnan et~al.(2023)Radhakrishnan, Ruiz~Luyten, Prasad, and Uhler]{Radhakrishnan2023-tg}
Adityanarayanan Radhakrishnan, Max Ruiz~Luyten, Neha Prasad, and Caroline Uhler.
\newblock Transfer learning with kernel methods.
\newblock \emph{Nature Communications}, 14\penalty0 (1):\penalty0 5570, September 2023.

\bibitem[Raghu et~al.(2019)Raghu, Zhang, Kleinberg, and Bengio]{DBLP:journals/corr/abs-1902-07208}
Maithra Raghu, Chiyuan Zhang, Jon~M. Kleinberg, and Samy Bengio.
\newblock Transfusion: Understanding transfer learning with applications to medical imaging.
\newblock \emph{CoRR}, abs/1902.07208, 2019.
\newblock URL \url{http://arxiv.org/abs/1902.07208}.

\bibitem[Scarselli et~al.(2009)Scarselli, Gori, Tsoi, Hagenbuchner, and Monfardini]{scarselli2009}
Franco Scarselli, Marco Gori, Ah~Tsoi, Markus Hagenbuchner, and Gabriele Monfardini.
\newblock The graph neural network model.
\newblock \emph{IEEE transactions on neural networks / a publication of the IEEE Neural Networks Council}, 20:\penalty0 61--80, 01 2009.
\newblock \doi{10.1109/TNN.2008.2005605}.

\bibitem[Sun et~al.(2024)Sun, Liao, MacDonald, Zhang, Liu, Huguet, Wolf, Adelstein, Rudner, and Krishnaswamy]{sun2024geometry}
Xingzhi Sun, Danqi Liao, Kincaid MacDonald, Yanlei Zhang, Chen Liu, Guillaume Huguet, Guy Wolf, Ian Adelstein, Tim~GJ Rudner, and Smita Krishnaswamy.
\newblock Geometry-aware generative autoencoders for warped riemannian metric learning and generative modeling on data manifolds.
\newblock \emph{CoRR}, 2024.

\bibitem[Sushko et~al.(2011)Sushko, Novotarskyi, K{\"o}rner, Pandey, Rupp, Teetz, Brandmaier, Abdelaziz, Prokopenko, Tanchuk, et~al.]{sushko2011online}
Iurii Sushko, Sergii Novotarskyi, Robert K{\"o}rner, Anil~Kumar Pandey, Matthias Rupp, Wolfram Teetz, Stefan Brandmaier, Ahmed Abdelaziz, Volodymyr~V Prokopenko, Vsevolod~Y Tanchuk, et~al.
\newblock Online chemical modeling environment (ochem): web platform for data storage, model development and publishing of chemical information.
\newblock \emph{Journal of computer-aided molecular design}, 25:\penalty0 533--554, 2011.

\bibitem[Wang et~al.(2019)Wang, Agarwal, Huang, Hu, Zhou, Ye, and Zhang]{wang2019}
Jingshu Wang, Divyansh Agarwal, Mo~Huang, Gang Hu, Zilu Zhou, Chengzhong Ye, and Nancy Zhang.
\newblock Data denoising with transfer learning in single-cell transcriptomics.
\newblock \emph{Nature Methods}, 16:\penalty0 875--878, 09 2019.
\newblock \doi{10.1038/s41592-019-0537-1}.

\bibitem[Wenzel et~al.(2022)Wenzel, Dittadi, Gehler, Simon-Gabriel, Horn, Zietlow, Kernert, Russell, Brox, Schiele, Schölkopf, and Locatello]{wenzel2022assayingoutofdistributiongeneralizationtransfer}
Florian Wenzel, Andrea Dittadi, Peter~Vincent Gehler, Carl-Johann Simon-Gabriel, Max Horn, Dominik Zietlow, David Kernert, Chris Russell, Thomas Brox, Bernt Schiele, Bernhard Schölkopf, and Francesco Locatello.
\newblock Assaying out-of-distribution generalization in transfer learning, 2022.
\newblock URL \url{https://arxiv.org/abs/2207.09239}.

\bibitem[Yang et~al.(2019)Yang, Swanson, Jin, Coley, Eiden, Gao, Guzman-Perez, Hopper, Kelley, Mathea, Palmer, Settels, Jaakkola, Jensen, and Barzilay]{dmpnn}
Kevin Yang, Kyle Swanson, Wengong Jin, Connor Coley, Philipp Eiden, Hua Gao, Angel Guzman-Perez, Tim Hopper, Brian Kelley, Miriam Mathea, Andrew Palmer, Volker Settels, Tommi Jaakkola, Klavs Jensen, and Regina Barzilay.
\newblock Analyzing learned molecular representations for property prediction.
\newblock \emph{Journal of Chemical Information and Modeling}, 59, 07 2019.
\newblock \doi{10.1021/acs.jcim.9b00237}.

\bibitem[Yang et~al.(2018)Yang, Arvanitidis, Fu, Li, and Hauberg]{yang2018geodesic}
Tao Yang, Georgios Arvanitidis, Dongmei Fu, Xiaogang Li, and S{\o}ren Hauberg.
\newblock Geodesic clustering in deep generative models.
\newblock \emph{arXiv preprint arXiv:1809.04747}, 2018.

\bibitem[Yao et~al.(2024)Yao, Song, Jia, Cheng, Zhong, Song, and Feng]{yao2024fast}
Shaolun Yao, Jie Song, Lingxiang Jia, Lechao Cheng, Zipeng Zhong, Mingli Song, and Zunlei Feng.
\newblock Fast and effective molecular property prediction with transferability map.
\newblock \emph{Communications Chemistry}, 7\penalty0 (1):\penalty0 85, 2024.

\bibitem[Yim et~al.(2024)Yim, Jeong, Ko, Lee, Kim, Lee, and Han]{yim2024taskadditionmultitasklearning}
Soorin Yim, Dae-Woong Jeong, Sung~Moon Ko, Sumin Lee, Hyunseung Kim, Chanhui Lee, and Sehui Han.
\newblock Task addition in multi-task learning by geometrical alignment, 2024.
\newblock URL \url{https://arxiv.org/abs/2409.16645}.

\bibitem[Yu et~al.(2022)Yu, Wang, Hong, Teku, Wang, and Zhang]{YU2022230}
Xiang Yu, Jian Wang, Qing-Qi Hong, Raja Teku, Shui-Hua Wang, and Yu-Dong Zhang.
\newblock Transfer learning for medical images analyses: A survey.
\newblock \emph{Neurocomputing}, 489:\penalty0 230--254, 2022.
\newblock ISSN 0925-2312.
\newblock \doi{https://doi.org/10.1016/j.neucom.2021.08.159}.
\newblock URL \url{https://www.sciencedirect.com/science/article/pii/S0925231222003174}.

\bibitem[Zhuang et~al.(2011)Zhuang, Luo, Xiong, He, Xiong, and Shi]{https://doi.org/10.1002/sam.10099}
Fuzhen Zhuang, Ping Luo, Hui Xiong, Qing He, Yuhong Xiong, and Zhongzhi Shi.
\newblock Exploiting associations between word clusters and document classes for cross-domain text categorization†.
\newblock \emph{Statistical Analysis and Data Mining: The ASA Data Science Journal}, 4\penalty0 (1):\penalty0 100--114, 2011.
\newblock \doi{https://doi.org/10.1002/sam.10099}.
\newblock URL \url{https://onlinelibrary.wiley.com/doi/abs/10.1002/sam.10099}.

\bibitem[Zhuang et~al.(2013)Zhuang, Luo, Du, He, and Shi]{10.1145/2433396.2433449}
Fuzhen Zhuang, Ping Luo, Changying Du, Qing He, and Zhongzhi Shi.
\newblock Triplex transfer learning: Exploiting both shared and distinct concepts for text classification.
\newblock In \emph{Proceedings of the Sixth ACM International Conference on Web Search and Data Mining}, WSDM '13, page 425–434, New York, NY, USA, 2013. Association for Computing Machinery.
\newblock ISBN 9781450318693.
\newblock \doi{10.1145/2433396.2433449}.
\newblock URL \url{https://doi.org/10.1145/2433396.2433449}.

\bibitem[Zhuang et~al.(2014)Zhuang, Luo, Du, He, Shi, and Xiong]{6606822}
Fuzhen Zhuang, Ping Luo, Changying Du, Qing He, Zhongzhi Shi, and Hui Xiong.
\newblock Triplex transfer learning: Exploiting both shared and distinct concepts for text classification.
\newblock \emph{IEEE Transactions on Cybernetics}, 44\penalty0 (7):\penalty0 1191--1203, 2014.
\newblock \doi{10.1109/TCYB.2013.2281451}.

\end{thebibliography}

\appendix

\newpage
\section{Notations}
\label{notations}
Our notation follows index notation and the Einstein summation convention. The functions and matrices used in our algorithm are defined as follows.
 \begin{eqnarray}
     X : \textrm{Vector} \\ 
     X^{\mu} : \textrm{Vector Field} \\ 
     dx_\mu : \textrm{Basis}\\
     X_{\mu} : \textrm{Dual Vector Field} \\ 
     dx^\mu : \textrm{Dual Basis}\\
     T : \textrm{Tensor} \\ 
     T^{\nu_1 \cdots \nu_p}_{\phantom{\nu_1 \cdots \nu_p}\mu_1 \cdots \mu_q} : \textrm{(p, q) Tensor Field} \\ 
     g_{\mu\nu} : \textrm{Metric Tensor}\\
     \delta_{\mu\nu} : \textrm{Kronecker Delta}\\
     \nabla_\mu : \textrm{Covariant Derivative}\\
     \mathcal{L}_X : \textrm{Lie Derivative}\\
     \Gamma^{\rho}_{\phantom{\rho}\mu\nu} : \textrm{Christoffel Symbol}
 \end{eqnarray}
 
 All indices are raised and lowered by the metric $g_{\mu\nu}$. For instances,
 
 \begin{equation}
     g^{\mu}_{\phantom{\mu}\nu} = g^{\mu\rho} g_{\rho\nu}
 \end{equation}

where

\begin{equation}
    g^{\mu\nu}g_{\mu\nu} = \delta^\mu_{\phantom{\mu}\nu} = D
\end{equation}

Here $D$ is the number of dimensions.

%%%%%%%%%%%%%%%%%%%%%%%%%%%%%%%%%%%%%%%%%%%%%%%%%%%%%%%%%%%%%%%%%%%%%%%%%%%%
%%%%%%%%%%%%%%%%%%%%%%%%%%%%%%%%%%%%%%%%%%%%%%%%%%%%%%%%%%%%%%%%%%%%%%%%%%%%
%%%%%%%%%%%%%%%%%%%%%%%%%%%%%%%%%%%%%%%%%%%%%%%%%%%%%%%%%%%%%%%%%%%%%%%%%%%%
\section{Proofs and Derivations}
\label{proofs}
\subsection{The Definition of Riemannian Manifold}
A curved space is complicated to comprehend in general. Since the late 19th century, there has been immense development in differential geometry to formally interpret curved spaces. One of the best-known intuitive geometries is Riemannian geometry. Riemannian geometry possesses a handful of useful mathematical properties that can be utilized in the real world. The formal definition of Riemannian geometry is as follows:
\begin{definition}[Riemannian Manifold]
A Riemannian metric on a smooth manifold M is a choice at each point $x \in M$ of a positive definite inner product $g_p: T_pM \times T_pM \rightarrow \mathbb{R}$ on $T_xM$. The smooth manifold endowed with the metric $g$ is a Riemannian manifold, denoted $(M, g)$.
\end{definition}
As stated above, a Riemannian manifold is smooth and differentiable everywhere on the manifold, along with its derivatives. Moreover, a Riemannian manifold enjoys diffeomorphism invariance, induced by the Lie derivative $\mathcal{L}_X$. It can be readily observed that the composition of two different Lie derivatives forms a group, known as the diffeomorphism group. This isometry guarantees that coordinate choices can be made without altering the global geometry of the space.
\begin{equation} \label{vec_trans}
    X' = X'^\mu dX'_\mu = X'^\mu \frac{\partial X^\nu}{\partial X'^\mu} dX_\nu = X^\nu dX_\nu = X
\end{equation}
As shown in Eq.~\ref{vec_trans}, the transformed vector remains unchanged. Moreover, it is always possible to fix the transformed coordinates in a locally flat space.
\begin{equation}
    \xi^\mu = \frac{\partial \xi^\mu}{\partial X^\nu} X^\nu
\end{equation}
Where $\xi^\mu$ is a vector on a locally flat frame. To ensure the vector is on a flat frame, one must impose the following condition:
\begin{equation}
    \frac{\partial^2}{\partial t^2}\xi^\mu (t) \equiv 0
\end{equation}
Since a vector is on a flat frame, it should be in free-falling motion, and thus its acceleration should be trivial. On a locally flat frame, the metric also reduces to the flat Euclidean metric.
\begin{equation}
    g_{\mu\nu} = 1_{\mu\nu}
\end{equation}
\subsection{Covariance}
A vector should transform consistently across any coordinate frame. However, if the space is no longer flat, the ordinary derivative no longer preserves this property. To address this, let us consider the derivative of a vector in a general curved space.
\begin{equation}
    \partial_\mu \rightarrow \partial'_\mu = \frac{\partial x^\mu}{\partial x'^\nu} \partial_\nu
\end{equation}
Where $\partial_\mu = \frac{\partial}{\partial x^\mu}$, the vector transformation can be written as follows:
\begin{eqnarray}
    \partial_\nu X^\mu \rightarrow \partial'_\nu X'^\mu &=& \frac{\partial x^\lambda}{\partial x'^\nu} \frac{\partial}{\partial x^\lambda}(\frac{\partial x'^\mu}{\partial x^\rho}V^\rho)\\
    &=& \frac{\partial x'^\nu}{\partial x^\lambda}\Big(\frac{\partial x'^\rho}{\partial x^\nu} \partial^\lambda V^\rho + \frac{\partial^2 x'^\mu}{\partial x^\lambda \partial x^\rho} V^\rho \Big)
\end{eqnarray}
As shown above, the transformation of a vector on a curved space using an ordinary derivative is no longer covariant. Therefore, it is necessary to introduce an additional term to restore covariance, namely the affine connection. With this addition, one can define the covariant derivative, which replaces the ordinary derivative.
\begin{equation}
    \nabla_\mu = \partial_\mu + \Gamma^\lambda_{\phantom{\lambda}\mu\nu}
\end{equation}
By imposing the covariance condition on the covariant derivative,
\begin{equation}
    \nabla_\lambda \rightarrow \nabla'_\lambda V'^\mu = \frac{\partial x^\rho}{\partial x'^\nu}\frac{\partial x'^\mu}{\partial x^\nu}\nabla_\rho V^\nu
\end{equation}
one can derive the explicit form of the connection.
\begin{equation}
    \nabla_\mu V^\nu = \partial_\mu V^\nu + \Gamma^\nu_{\phantom{\nu}\mu\lambda} V^\lambda
\end{equation}
Under coordinate transformation,
\begin{equation}
    \frac{\partial}{\partial x'^{\mu}}(\frac{\partial x'^\nu}{\partial x^\lambda}V^\lambda) + \Gamma'^\nu_{\phantom{\nu}\mu\sigma}V'^\sigma = \frac{\partial x^\rho}{\partial x'^\mu} \frac{\partial x'^\nu}{\partial x^\lambda} \partial_\rho V^\lambda + \frac{\partial x^\rho}{\partial x'^\mu} \frac{\partial^2 x'^\nu}{\partial x^\rho \partial x^\lambda} V^\lambda + \Gamma'^\nu_{\phantom{\nu}\mu\sigma} V'^\sigma
\end{equation}
Here, to make the derivative of a vector covariant, the following condition must be satisfied:
\begin{equation}
    \frac{\partial x^\rho}{\partial x'^\mu} \frac{\partial^2 x'^\nu}{\partial x^\rho \partial x^\lambda} V^\lambda + \Gamma'^\nu_{\phantom{\nu}\mu\sigma}V'^\sigma = \frac{\partial x^\rho}{\partial x'^\mu} \frac{\partial x'^\nu}{\partial x^\lambda} \Gamma^\lambda_{\phantom{\lambda}\rho\sigma}V^\sigma
\end{equation}
Which is
\begin{gather}
    \Gamma'^\nu_{\phantom{\nu}\mu\sigma}(\frac{\partial x'^\sigma}{\partial x^\tau}V^\tau) = \frac{\partial x^\rho}{\partial x'^\mu} \frac{\partial'^\nu}{\partial x^\lambda}\Gamma^\lambda_{\phantom{\lambda}\rho\sigma}V^\sigma - \frac{\partial x^\rho}{\partial x'^\mu} \frac{\partial x^\rho}{\partial x'^\mu} \frac{\partial^2 x'^\nu}{\partial x^\rho \partial x^\lambda} V^\lambda\\
    \Gamma'^\nu_{\phantom{\nu}\mu\kappa} V^\tau = \frac{\partial x^\rho}{\partial x'^\kappa} \frac{\partial x^\rho}{\partial x'^\mu} \frac{\partial x'^\nu}{\partial x^\lambda} \Gamma^\lambda_{\phantom{\lambda}\rho\sigma} V^\sigma - \frac{\partial x^\tau}{\partial x'^\kappa} \frac{\partial x^\rho}{\partial x'^\mu} \frac{\partial^2 x'^\nu}{\partial x^\rho \partial x^\lambda} V^\lambda
\end{gather}
This leads to the explicit form of how the Christoffel symbols transform under coordinate changes.
\begin{gather}
    \Gamma'^\nu_{\phantom{\nu}\mu\kappa} = \frac{\partial x^\tau}{\partial x'^\kappa} \frac{\partial x^\rho}{\partial x'^\mu} \frac{\partial x'^\nu}{\partial x^\lambda} \Gamma^\lambda_{\phantom{\lambda}\rho\tau} - \frac{\partial x^\tau}{\partial x'^\kappa} \frac{\partial x^\rho}{\partial x'^\mu} \frac{\partial^2 x'^\nu}{\partial x^\rho \partial x^\tau}
\end{gather}
Since the Kronecker delta is a constant matrix, it is clear that its derivative must vanish. By applying the chain rule to the delta, one can derive the following relation, which simplifies the transformation rule described above.
\begin{equation}
    \frac{\partial}{\partial x'^\mu} \delta^\nu_\kappa = \frac{\partial}{\partial x'^\mu} \frac{\partial x'^\nu}{\partial x'^\kappa} = \frac{\partial}{\partial x'^\mu}(\frac{\partial x^\tau}{\partial x'^\kappa} \frac{\partial x'^\nu}{\partial x^\tau}) = 0 = \frac{\partial x^\tau}{\partial x'^\kappa}\frac{\partial x^\rho}{\partial x'^\mu} \frac{\partial^2 x'^\nu}{\partial x^\rho \partial x^\tau} + \frac{\partial x'^\nu}{\partial x^\tau} \frac{\partial x'^\nu}{\partial x^\tau} \frac{\partial^2 x^\tau}{\partial x'^\mu \partial x'^\rho}
\end{equation}
Finally, the transformation rule for the Christoffel symbols is given by:
\begin{equation}
    \Gamma'^\nu_{\phantom{\nu}\mu\kappa} = \frac{\partial x^\tau}{\partial x'^\kappa} \frac{\partial x^\rho}{\partial x'^\mu} \frac{\partial x'^\nu}{\partial x^\lambda} \Gamma^\lambda_{\phantom{\lambda}\rho\tau} + \frac{\partial x'^\nu}{\partial x^\tau} \frac{\partial^2 x^\tau}{\partial x'^\mu \partial x'^\rho}
\end{equation}
By the same reasoning, one can easily determine how covariant derivatives act on differential forms.
\begin{equation}
    \nabla_\mu V_\nu = \partial_\mu V_\nu - \Gamma^\lambda_{\phantom{\lambda}\mu\nu}V_\lambda
\end{equation}

\subsection{Explicit Form of Christoffel Symbol}
The metric serves as the ruler of a given geometry; therefore, it should remain invariant with respect to position in a coordinate system. In the case of Euclidean space, this invariance is trivial to observe, as the metric is simply $\delta_{\mu\nu}$, a constant matrix.
\begin{equation}
    \frac{\partial}{\partial x^\lambda} \delta_{\mu\nu} = 0
\end{equation}
However, in the curved case, the above principle must still hold to interpret the metric as a ruler. Nevertheless, this condition does not hold when using an ordinary derivative. Here, the covariant derivative comes into play, replacing the ordinary derivative. When taking the covariant derivative of the curved metric, the resulting term vanishes.
\begin{equation} \label{der_g}
    \nabla_\lambda g_{\mu\nu} = 0
\end{equation}
One can express this condition in terms of a flat metric combined with a diffeomorphism transformation factor.
\begin{equation}
    g_{\mu\nu}(x) = \frac{\partial \xi^\lambda}{\partial x^\mu} \frac{\partial \xi^\rho}{\partial x^\nu} \delta_{\lambda\rho}(\xi)
\end{equation}
Taking the derivative with respect to $x$ on both sides, the equation becomes:
\begin{eqnarray}
    \frac{\partial}{\partial x^\sigma} g_{\mu\nu}(x) & = & \frac{\partial^2 x^\lambda}{\partial x^\sigma \partial x^\mu} \frac{\xi^\rho}{\partial x^\nu} \delta_{\lambda \rho} + \frac{\partial^2 \xi^\rho}{\partial x^\sigma \partial x^\nu} \frac{\partial \xi^\lambda}{\partial x^\mu} \delta_{\lambda \rho}\\
    & = & \frac{\partial^2 \xi^\rho}{\partial x^\sigma \partial x^\nu} \frac{\partial x^\tau}{\partial \xi^\rho} \frac{\partial \xi^\rho}{\partial x^\tau}\frac{\partial \xi^\lambda}{\partial x^\mu} \delta_{\lambda \rho} + \frac{\partial^2 \xi^\lambda}{\partial x^\sigma \partial x^\mu} \frac{\partial x^\tau}{\partial \xi^\lambda} \frac{\partial \xi^\lambda}{\partial x^\tau}\frac{\partial \xi^\rho}{\partial x^\nu} \delta_{\lambda \rho}\\
    & = & \frac{\partial^2 \xi^\rho}{\partial x^\sigma \partial x^\nu} \frac{\partial x^\tau}{\partial \xi^\rho} g_{\mu\tau} + \frac{\partial^2 \xi^\lambda}{\partial x^\sigma \partial x^\mu} \frac{\partial x^\tau}{\partial \xi^\lambda} g_{\tau \nu}
\end{eqnarray}
From Eq.~\ref{der_g}, one can easily derive the explicit form of the Christoffel symbol in terms of the derivatives of the curved and flat coordinates.
\begin{gather}
    \frac{\partial}{\partial x^\sigma} g_{\mu\nu} = \Gamma^\tau_{\phantom{\tau}\sigma\mu} g_{\tau \nu} + \Gamma^\tau_{\phantom{\tau}\nu\sigma} g_{\mu \sigma} \\
    \Gamma^\tau_{\phantom{\tau}\sigma\mu} = \frac{\partial^2 \xi^\lambda}{\partial x^\sigma \partial x^\mu} \frac{\partial x^\tau}{\partial \xi^\lambda}(x)
\end{gather}
Since the metric should always be symmetric, the lower indices of the Christoffel symbol should also be symmetric. It is called a torsion-free condition. Furthermore, by utilizing a simple mathematical trick, one can obtain the Christoffel symbol in terms of the metric $g_{\mu\nu}$.
\begin{eqnarray}
\frac{\partial}{\partial x^\sigma} g_{\mu\nu} & = & \Gamma^\tau_{\phantom{\tau}\sigma\mu}g_{\tau\nu} +
\Gamma^\tau_{\phantom{\tau}\sigma\nu}g_{\mu\tau}\\
\frac{\partial}{\partial x^\mu} g_{\nu\sigma} & = & \Gamma^\tau_{\phantom{\tau}\mu\nu}g_{\tau\sigma} +
\Gamma^\tau_{\phantom{\tau}\mu\sigma}g_{\nu\tau}\\
\frac{\partial}{\partial x^\nu} g_{\sigma\mu} & = & \Gamma^\tau_{\phantom{\tau}\nu\sigma}g_{\tau\mu} +
\Gamma^\tau_{\phantom{\tau}\nu\mu}g_{\sigma\tau}
\end{eqnarray}
Adding the first two equations and subtracting the last one leads to:
\begin{equation}
    \Gamma^\lambda_{\phantom{\lambda}\mu\nu} = \frac{1}{2} g^{\lambda \rho}(\frac{\partial}{\partial x^\mu}g_{\nu\rho} + \frac{\partial}{\partial x^\nu}g_{\rho\mu} - \frac{\partial}{\partial x^\rho}g_{\mu\nu})
\end{equation}

\subsection{Geodesic Equations}
The shortest path between two points is simple to define in flat space. However, in curved space, this notion becomes more complicated. The shortest path in a curved space is defined as a geodesic. There are several ways to derive the geodesic equation, one of which is by imposing the free-fall condition.
\begin{equation}
    \frac{\partial^2 \xi^\mu (\tau)}{\partial \tau^2} = 0
\end{equation}
By a diffeomorphism, one can transform a coordinate into an arbitrary coordinate $x$.
\begin{gather}
0 = \frac{\partial}{\partial \tau}(\frac{\partial \xi^\mu}{\partial x^\nu} \frac{\partial x^\nu}{\partial \tau}) = \frac{\partial \xi ^\mu}{\partial x^\nu} \frac{\partial^2 x^\nu}{\partial \tau^2} + \frac{\partial^2 \xi^\mu}{\partial x^\lambda \partial x^\nu} \frac{\partial x^\lambda}{\partial \tau} \frac{\partial x^\nu}{\partial \tau}\\
\frac{\partial^2 x^\rho}{\partial \tau^2} + \frac{\partial^2 \xi^\mu}{\partial x^\lambda \partial x^\nu} \frac{\partial x^\rho}{\partial \xi^\mu} \frac{\partial x^\lambda}{\partial \tau} \frac{\partial x^\nu}{\partial \tau} = \frac{\partial^2 x^\rho}{\partial \tau^2} + \Gamma^\rho_{\phantom{\rho}\lambda\nu} \frac{\partial x^\lambda}{\partial \tau} \frac{\partial x^\nu}{\partial \tau} = 0
\end{gather}
Another way to derive the equation is by minimizing the distance in curved space.
\begin{equation}
    S = \int \sqrt{g_{\mu\nu} \frac{dx^\mu}{d\tau} \frac{dx^\nu}{d\tau}}d\tau
\end{equation}
By varying the above equation and requiring the variation to vanish, one can compute its minimum value, and after some tedious calculations, the geodesic equation can be obtained.
%%%%%%%%%%%%%%%%%%%%%%%%%%%%%%%%%%%%%%%%%%%%%%%%%%%%%%%%%%%%%%%%%%%%%%%%%%%%
%%%%%%%%%%%%%%%%%%%%%%%%%%%%%%%%%%%%%%%%%%%%%%%%%%%%%%%%%%%%%%%%%%%%%%%%%%%%
%%%%%%%%%%%%%%%%%%%%%%%%%%%%%%%%%%%%%%%%%%%%%%%%%%%%%%%%%%%%%%%%%%%%%%%%%%%%

\subsection{Riemann Curvature}
The Riemann curvature can be defined through the concept of parallel transport of a vector. In flat space, a vector remains unchanged under parallel transport along any path. However, in curved space, the vector's outcome depends on the path taken. This leads to the idea of curvature as the difference between the results of transporting a vector along two different paths from the same starting point to the same endpoint. This difference quantitatively characterizes the curvature of the space.
\begin{equation}
    [\nabla_\mu, \nabla_\nu]V^\lambda
\end{equation}
Here, the bracket denotes the commutation relation between the entities. Since the covariant derivative acts as the generator of parallel transport, the equation can be interpreted as the vector $V^\lambda$ being transported along two different paths: one generated by applying $\nabla_\mu$ followed by $\nabla_\nu$, and the other by reversing the order. The resulting computation takes the form:
\begin{equation}
    \begin{array}{rcl}
         \nabla_\mu \nabla_\nu V^\lambda & = & \partial_\mu (\nabla_\nu V^\lambda) + \Gamma^\lambda_{\phantom{\lambda}\mu\rho}\nabla_\nu V^\rho - \Gamma^\rho_{\phantom{\rho}\mu\nu} \nabla_\rho V^\lambda  \\
         & = & \partial_\mu \partial_\nu V^\lambda + \partial_\mu \Gamma^\lambda_{\phantom{\lambda}\nu\rho}V^\rho + \Gamma^\lambda_{\phantom{\lambda}\nu\rho} \partial_\mu V^\rho + \Gamma^\lambda_{\phantom{\lambda}_\mu\rho}\partial_\nu V^\rho\\
         &&+\Gamma^\lambda_{\phantom{\lambda}\mu\rho} \Gamma^\rho_{\phantom{\rho}\nu\sigma}V^\sigma - \Gamma^\rho_{\phantom{\rho}\mu\nu}\partial_\rho V^\lambda - \Gamma^\rho_{\phantom{\rho}_\mu\nu} \Gamma^\lambda_{\phantom{\lambda}\rho\sigma} V^\sigma
    \end{array}
\end{equation}
Where,
\begin{equation}
    [\nabla_\mu, \nabla_\nu]V^\lambda = (\partial_\mu \Gamma^\lambda_{\phantom{\lambda}\nu\rho} - \partial_\nu \Gamma^\lambda_{\phantom{\lambda}\mu\rho} + \Gamma^\lambda_{\phantom{\lambda}\mu\sigma} \Gamma^\sigma_{\phantom{\sigma}\nu\rho} - \Gamma^\lambda_{\phantom{\lambda}\nu\sigma}\Gamma^\sigma_{\phantom{\sigma}\mu\rho})V^\rho - 2\Gamma^\rho_{[\mu\nu]}\nabla_\rho V^\lambda
\end{equation}
Since the connection is symmetric under the permutation of its lower indices, the last term in the above equation can be eliminated. We can then finally define the Riemann tensor.
\begin{equation} \label{Riemman}
    R^\lambda_{\phantom{\lambda}\rho\mu\nu} := \partial_\mu \Gamma^\lambda_{\phantom{\lambda}\nu\rho} - \partial_\nu \Gamma^\lambda_{\phantom{\lambda}\mu\rho} + \Gamma^\lambda_{\phantom{\lambda}\mu\sigma} \Gamma^\sigma_{\phantom{\sigma}\nu\rho} - \Gamma^\lambda_{\phantom{\lambda}\nu\sigma}\Gamma^\sigma_{\phantom{\sigma}\mu\rho}
\end{equation}
The Riemann curvature tensor possesses several useful properties.
\begin{equation}
\begin{array}{l}
     R_{\lambda\rho\mu\nu} = -R_{\lambda\rho\nu\mu} \\
     R_{\lambda\rho\mu\nu} = -R_{\rho\lambda\nu\mu} \\
     R_{\lambda\rho\mu\nu} = -R_{\nu\mu\lambda\rho} \\
     R_{\lambda\rho\mu\nu} + R_{\lambda\mu\nu\rho} + R_{\lambda\nu\rho\mu} = 0\\
     \nabla_\sigma R_{\lambda\rho\mu\nu} + \nabla_\mu R_{\lambda\rho\nu\sigma} + \nabla_\nu R_{\lambda\rho\sigma\mu} = 0 \\
     \nabla_\sigma R_{\lambda\rho\mu\nu} + \nabla_\lambda R_{\rho\sigma\nu\mu} + \nabla_\rho R_{\sigma\lambda\mu\nu} = 0 \\
     R_{\lambda\rho\mu\nu} + R_{\rho\mu\lambda\nu} + R_{\mu\lambda\rho\nu} = 0
      
\end{array}
\end{equation}
With the curved metric $g_{\mu\nu}$, one can construct the Ricci curvature tensor and the Ricci scalar by contracting the first and third, and the second and fourth indices of the Riemann tensor, respectively.
\begin{equation}
\begin{array}{l} \label{curvatures}
     R_{\rho\nu} = g^{\lambda\mu} R_{\lambda\rho\mu\nu}\\
     R = g^{\rho\nu} R_{\rho\nu}  = g^{\rho\nu} g^{\lambda\mu} R_{\lambda\rho\mu\nu}
\end{array}
\end{equation}

%%%%%%%%%%%%%%%%%%%%%%%%%%%%%%%%%%%%%%%%%%%%%%%%%%%%%%%%%%%%%%%%%%%%%%%%%%%%
%%%%%%%%%%%%%%%%%%%%%%%%%%%%%%%%%%%%%%%%%%%%%%%%%%%%%%%%%%%%%%%%%%%%%%%%%%%%
%%%%%%%%%%%%%%%%%%%%%%%%%%%%%%%%%%%%%%%%%%%%%%%%%%%%%%%%%%%%%%%%%%%%%%%%%%%%

\section{Curvature Computation for Two Layered MLP}\label{curvature computation}
Before we proceed, it is convenient to define some symbols that will frequently appear in the following calculations. Since we will use the SiLU activation function, the logistic function will appear repeatedly in derivative computations. Therefore, we introduce the following symbol to represent the logistic function:
\begin{equation}
    \logi(x) \equiv \frac{1}{1+e^{-x}}
\end{equation}
The derivative of the logistic function is well known. It consists of the square of the logistic function multiplied by an $x e^{-x}$ term.
\begin{equation}
    \frac{d}{dx} \logi(x) = e^{-x} \times \logi(x)^2 = \frac{e^{-x}}{(1+e^{-x})^2}
\end{equation}
The coordinate transformation function in the model is based on an MLP with the SiLU activation function. To obtain the Jacobian of the transformation function, one must differentiate the transformation function with respect to the transformed coordinate $x'$.
\begin{equation}
    J^{i}_{\phantom{i}j} = \frac{dx'^i}{dx^j}
\end{equation}
Here, $x'^i$ can be expressed in the following form.
\begin{equation}
    x'^i = W^{(2)i}_{\phantom{(1)i}k}f(W^{(1)k}_{\phantom{(1)k}j}x^j + b^{(1)k}) + b^{(2)i}
\end{equation}
$W^{(n)i}_{\phantom{(n)i}_j}$, $b^{(n)i}$, and $f(x)$ denote the weight matrix, bias for each distinct hidden layer, and activation function, respectively. The activation function is, in this case, SiLU. Thus, the derivative of the function becomes straightforward.
\begin{equation} \label{jaco}
    \frac{dx'^i}{dx^j} = W^{(2)i}_{\phantom{(2)i}m} W^{(1)m}_{\phantom{(1)m}k} f(W^{k}_{\phantom{i}l}x^l + b^k)_{,j}
\end{equation}
The derivative of the activation term is tedious but manageable. First, we will show how the derivative of the SiLU function appears.
\begin{equation}
    \textrm{SiLU}(x) \equiv x \times \logi(x)
\end{equation}
Hence, the derivative can be expressed as follows:
\begin{equation} \label{dsilu}
    \frac{d}{dx}\textrm{Silu}(x) = x\times \logi(x)' + \logi(x) = \frac{xe^{-x}}{(1+e^{-x})^2} + \frac{1}{1+e^{-x}}
\end{equation}
For computational convenience, we first derive the derivative of the exponential term with respect to the weight and bias.
\begin{equation}\label{dE}
    \partial_k e^{-(W^i_{\phantom{i}j}x^j + b^i)} = -W^l_{\phantom{l}k}(e^{W^l_{\phantom{l}j}x^j + b^l})^{i}_{\phantom{i}l} = -W^l_{\phantom{l}k}E^i_{\phantom{i}l}
\end{equation}
Where $(e^{-(W^l_{\phantom{l}j}x^j + b^l)})^{i}_{\phantom{i}l}$ is a diagonal form as follows.
\begin{equation}
\textrm{E}^{i}_{\phantom{i}l} \equiv (e^{-(W^l_{\phantom{l}j}x^j + b^j)})^{i}_{\phantom{i}l} = \Bigg\{
\begin{array}{cl}
     e^{-(W^l_{\phantom{l}j}x^j + b^l)} & \textrm{if $l = i$}\\
      0   & \textrm{if $l \neq i$}
\end{array}
\end{equation}
By plugging Eq.\ref{dsilu} into Eq.\ref{jaco}, one can obtain the final form of the Jacobian. To express the equation in a simpler form, it is convenient to introduce the following symbols before proceeding with the main computation.
\begin{eqnarray}
    \sigma^i = \frac{1}{1+e^{-(W^{(1)i}_{\phantom{(1)i}j} x^j + b^{(1)i})}} \\
    \partial_j \sigma^i = W^{(1)l}_{\phantom{(1)l}j}E^i_{\phantom{i}l} (\sigma^2)^i\\
    x'^i = W^{(1)i}_{\phantom{(1)i}j} x^j + b^{(1)i} = (W^{(1)}x+b^{(1)})^i
\end{eqnarray}
Then, the Jacobian can be expressed in terms of the symbols introduced above.
\begin{eqnarray} \label{jaco2}
\begin{array}{rl}
\frac{dx'^i}{dx^j} =& W^{(2)i}_{\phantom{(2)i}m} W^{(1)m}_{\phantom{(1)m}k} (((x^{(1)k})e^{-x^{(1)k}} \times \logi(x^{(1)k}) + 1)\logi(x^{(1)k}))^k_{\phantom{k}j} \\
    =& \sum_{a_1} W^{(2)i}_{\phantom{(2)i}a_1} (W^{(1)a_1}_{\phantom{(1)a_1}j}\sigma^{a_1} + (W^{(1)}x+b^{(1)})^{a_1}W^{(1)a_3}_{\phantom{(1)a_3}j}E^{a_1}_{\phantom{a_1}a_3}(\sigma^2)^{a_1})
\end{array}
\end{eqnarray}
Due to the diffeomorphism invariance of a Riemannian manifold, the metric tensor can be decomposed into the square of the Jacobian of the given coordinate transformation, coupled with vectors and the flat Euclidean metric.
\begin{equation}
    g_{ij} = \frac{dx'^m}{dx^i} \frac{dx'^n}{dx^j} \eta_{mn} = \frac{dx'^m}{dx^i} \frac{dx'_m}{dx^j}
\end{equation}
where the metric can be explicitly expressed using Eq.~\ref{jaco}.
\begin{equation}
\begin{array}{rl} \label{metric}
    g_{ij} =&  \sum_{a_1 a_6} W^{(2)a_4}_{\phantom{(2)a_4}a_1} (W^{(1)a_1}_{\phantom{(1)a_1}i}\sigma^{a_1} + (W^{(1)}x+b^{(1)})^{a_1}W^{(1)a_3}_{\phantom{(1)a_3}i}E^{a_1}_{\phantom{a_1}a_3}(\sigma^2)^{a_1}) \\
    &W^{(2)}_{\phantom{(2)}a_4a_6} (W^{(1)a_6}_{\phantom{(1)a_6}j}\sigma^{a_6} + (W^{(1)}x+b^{(1)})^{a_6}W^{(1)a_7}_{\phantom{(1)a_7}j}E^{a_6}_{\phantom{a_6}a_7}(\sigma^2)^{a_6})
\end{array}
\end{equation}
As shown above, the curved metric $g_{ij}$ can be expressed in terms of the weight, bias, and input vector $x$. By taking derivatives and appropriately contracting the metric, the curvature tensor can also be expressed in terms of these components. The curvature tensor consists of combinations of derivatives of the Christoffel symbols. To compute a Christoffel symbol, one must first obtain the derivative of the metric tensor. The derivative of the metric tensor can be expressed as follows:
\begin{equation}
    \partial_k g_{ij} = \Big(\frac{\partial^2 x'^m}{\partial x^k \partial x^i} \frac{\partial x'^n}{\partial x^j} + \frac{\partial^2 x'^n}{\partial x^k \partial x^j} \frac{\partial x'^m}{\partial x^i} \Big) \eta_{mn}
\end{equation}
Here, the key term is the second derivative of a vector. To compute this second derivative, one must consider the derivative of the $\textrm{E}^{ij}$ term.
\begin{equation}\label{ddE}
    \partial_k \textrm{E}^{i}_{\phantom{i}j} = -W^l_{\phantom{l}k} \delta^i_{\phantom{i}lp}\textrm{E}^{p}_{\phantom{p}j}
\end{equation}
By utilizing the relation above, one can compute the second derivative of an arbitrary vector $x'^i$, which is a crucial component for deriving the affine connection.
\begin{equation}
\begin{array}{ll}
    \frac{\partial^2 x'^i}{\partial x^k \partial x^j} =& \sum_{a_1}W^{(2)i}_{\phantom{(2)i}a_1}(W^{(1)a_1}_{\phantom{(1)a_1}j}W^{(1)a_2}_{\phantom{(1)a_2}k}E^{a_1}_{\phantom{a_1}a_2}(\sigma^2)^{a_1}+ W^{(1)a_1}_{\phantom{(1)a_1}k}W^{(1)a_3}_{\phantom{(1)a_3}j}E^{a_1}_{\phantom{a_1}a_3}(\sigma^2)^{a_1}\\
    & - (W^{(1)}x + b^{(1)})^{a_1}W^{(1)a_3}_{\phantom{(1)a_3}j}W^{(1)a_4}_{\phantom{(1)a_4}k}\delta^{a_1}_{\phantom{a_1}a_4 a_5}E^{a_5}_{\phantom{a_5}a_3}(\sigma^2)^{a_1}\\
    & + 2(W^{(1)}x + b^{(1)})^{a_1}W^{(1)a_3}_{\phantom{(1)a_3}j}E^{a_1}_{\phantom{a_1}a_3}W^{(1)a_6}_{\phantom{(1)a_6}k}E^{a_1}_{\phantom{a_1}a_6}(\sigma^3)^{a_1})
\end{array}
\end{equation}
Now, with the second derivative, we can construct the derivative of the metric.
\begin{equation}\label{dg}
\begin{array}{ll}
    \partial_k g_{ij} =& \sum_{a_1 a_7}(W^{(2)m}_{\phantom{(2)m}a_1}(W^{(1)a_1}_{\phantom{(1)a_1}i}W^{(1)a_2}_{\phantom{(1)a_2}k}E^{a_1}_{\phantom{a_1}a_2}(\sigma^2)^{a_1}+ W^{(1)a_1}_{\phantom{(1)a_1}k}W^{(1)a_3}_{\phantom{(1)a_3}i}E^{a_1}_{\phantom{a_1}a_3}(\sigma^2)^{a_1}\\
    & - (W^{(1)}x + b^{(1)})^{a_1}W^{(1)a_3}_{\phantom{(1)a_3}i}W^{(1)a_4}_{\phantom{(1)a_4}k}\delta^{a_1}_{\phantom{a_1}a_4 a_5}E^{a_5}_{\phantom{a_5}a_3}(\sigma^2)^{a_1}\\
    & + 2(W^{(1)}x + b^{(1)})^{a_1}W^{(1)a_3}_{\phantom{(1)a_3}i}E^{a_1}_{\phantom{a_1}a_3}W^{(1)a_6}_{\phantom{(1)a_6}k}E^{a_1}_{\phantom{a_1}a_6}(\sigma^3)^{a_1})\\
    &W^{(2)n}_{\phantom{(2)n}a_7} (W^{(1)a_7}_{\phantom{(1)a_7}j}\sigma^{a_7} + (W^{(1)}x+b^{(1)})^{a_7}W^{(1)a_8}_{\phantom{(1)a_8}j}E^{a_7}_{\phantom{a_7}a_8}(\sigma^2)^{a_7})\\
    & + W^{(2)n}_{\phantom{(2)n}a_1}(W^{(1)a_1}_{\phantom{(1)a_1}j}W^{(1)a_2}_{\phantom{(1)a_2}k}E^{a_1}_{\phantom{a_1}a_2}(\sigma^2)^{a_1}+ W^{(1)a_1}_{\phantom{(1)a_1}k}W^{(1)a_3}_{\phantom{(1)a_3}j}E^{a_1}_{\phantom{a_1}a_3}(\sigma^2)^{a_1}\\
    & - (W^{(1)}x + b^{(1)})^{a_1}W^{(1)a_3}_{\phantom{(1)a_3}j}W^{(1)a_4}_{\phantom{(1)a_4}k}\delta^{a_1}_{\phantom{a_1}a_4 a_5}E^{a_5}_{\phantom{a_5}a_3}(\sigma^2)^{a_1}\\
    & + 2(W^{(1)}x + b^{(1)})^{a_1}W^{(1)a_3}_{\phantom{(1)a_3}j}E^{a_1}_{\phantom{a_1}a_3}W^{(1)a_6}_{\phantom{(1)a_6}k}E^{a_1}_{\phantom{a_1}a_6}(\sigma^3)^{a_1})\\
    &
    W^{(2)m}_{\phantom{(2)m}a_7} (W^{(1)a_7}_{\phantom{(1)a_7}i}\sigma^{a_7} + (W^{(1)}x+b^{(1)})^{a_7}W^{(1)a_8}_{\phantom{(1)a_8}i}E^{a_7}_{\phantom{a_7}a_8}(\sigma^2)^{a_7}))\eta_{mn}
\end{array}
\end{equation}
Since the Christoffel symbol can be written in terms of the derivative of the given metric, one can now express the complete form of the symbol using Eq.~\ref{dg}.
\begin{equation}\label{connection}
    \Gamma^i_{\phantom{i}jk} = \frac{1}{2}g^{im}(\partial_j g_{mk} + \partial_k g_{mj} - \partial_m g_{kj})
\end{equation}
Here, $g^{im}$ is the inverse of the metric tensor, which satisfies the following relation:
\begin{eqnarray}
    g^{ij}g_{jk} = \delta^i_{\phantom{i}k}\\
    g^{ij}g_{ji} = D
\end{eqnarray}
where $D$ is the number of spatial dimensions. The inverse of the metric can be explicitly written using the inverse Jacobian.
\begin{equation}
    g^{ij} = \frac{dx^i}{dx'^m} \frac{dx^j}{dx'^n} \eta^{mn}
\end{equation}
Although the explicit form of the inverse metric is written in terms of combinations of inverse Jacobians, we design two distinct models to encapsulate the Jacobian and the inverse Jacobian separately for each case. Thus, the inverse Jacobian is not the actual matrix inverse of the Jacobian, but instead follows the same computational process as the Jacobian, with the model replaced by the inverse transfer model. Using this setup, one can then express the explicit form of the inverse metric tensor in terms of weights and biases.
\begin{equation}
\begin{array}{rl} \label{inv_metric}
    g^{ij} =&  \sum_{a_1 a_6} W'^{(2)a_4}_{\phantom{(2)a_4}a_1} (W'^{(1)a_1i}\sigma^{a_1} + (W'^{(1)}x+b^{(1)})^{a_1}W'^{(1)a_3i}E^{a_1}_{\phantom{a_1}a_3}(\sigma^2)^{a_1}) \\
    &W'^{(2)}_{\phantom{(2)}a_4a_6} (W'^{(1)a_6j}\sigma^{a_6} + (W'^{(1)}x+b^{(1)})^{a_6}W'^{(1)a_7j}E^{a_6}_{\phantom{a_6}a_7}(\sigma^2)^{a_6})
\end{array}
\end{equation}
The primed weights and biases indicate the weights and biases from the inverse transfer model. Finally, all individual components are now prepared to complete the expression for the Christoffel symbol. We now recall Eq.~\ref{connection}.
\begin{equation} \label{Gamma}
\begin{array}{rl}
\Gamma^i_{\phantom{i}jk} =& \frac{1}{2}g^{im}(\partial_j g_{mk} + \partial_k g_{mj} - \partial_m g_{kj})\\
=& \frac{1}{2} \sum_{a_1 a_4} W'^{(2)a_3}_{\phantom{(2)a_3}a_1} (W'^{(1)a_1i}\sigma^{a_1} + (W'^{(1)}x+b^{(1)})^{a_1}W'^{(1)a_2i}E^{a_1}_{\phantom{a_1}a_2}(\sigma^2)^{a_1}) \\
    &W'^{(2)}_{\phantom{(2)}a_3a_4} (W'^{(1)a_4m}\sigma^{a_4} + (W'^{(1)}x+b^{(1)})^{a_4}W'^{(1)a_5m}E^{a_4}_{\phantom{a_4}a_5}(\sigma^2)^{a_4})\\
    %%%%%%%%%%%%%%%%%%%%%%%%%%%%%%%%%%%%%%%%
    &\big(\sum_{a_5 a_{11}}W^{(2)o}_{\phantom{(2)o}a_5}((W^{(1)a_5}_{\phantom{(1)a_5}m}W^{(1)a_6}_{\phantom{(1)a_6}j}E^{a_5}_{\phantom{a_5}a_6}(\sigma^2)^{a_5}+ W^{(1)a_5}_{\phantom{(1)a_5}j}W^{(1)a_7}_{\phantom{(1)a_7}m}E^{a_5}_{\phantom{a_5}a_7}(\sigma^2)^{a_5}\\
    & - (W^{(1)}x + b^{(1)})^{a_5}W^{(1)a_7}_{\phantom{(1)a_7}m}W^{(1)a_8}_{\phantom{(1)a_8}j}\delta^{a_5}_{\phantom{a_5}a_8 a_9}E^{a_9}_{\phantom{a_9}a_7}(\sigma^2)^{a_5}\\
    & + 2(W^{(1)}x + b^{(1)})^{a_5}W^{(1)a_7}_{\phantom{(1)a_7}m}E^{a_5}_{\phantom{a_5}a_7}W^{(1)a_{10}}_{\phantom{(1)a{10}}j}E^{a_5}_{\phantom{a_5}a_{10}}(\sigma^3)^{a_5})\\
    &W^{(2)p}_{\phantom{(2)p}a_{11}} (W^{(1)a_{11}}_{\phantom{(1)a_{11}}k}\sigma^{a_{11}} + (W^{(1)}x+b^{(1)})^{a_{11}}W^{(1)a_{12}}_{\phantom{(1)a_{12}}k}E^{a_{11}}_{\phantom{a_{11}}a_{12}}(\sigma^2)^{a_{11}})\\
    & + W^{(2)p}_{\phantom{(2)p}a_5}(W^{(1)a_5}_{\phantom{(1)a_5}k}W^{(1)a_6}_{\phantom{(1)a_6}j}E^{a_5}_{\phantom{a_5}a_6}(\sigma^2)^{a_5}+ W^{(1)a_5}_{\phantom{(1)a_5}j}W^{(1)a_7}_{\phantom{(1)a_7}k}E^{a_5}_{\phantom{a_5}a_7}(\sigma^2)^{a_5}\\
    & - (W^{(1)}x + b^{(1)})^{a_5}W^{(1)a_7}_{\phantom{(1)a_7}k}W^{(1)a_8}_{\phantom{(1)a_8}j}\delta^{a_5}_{\phantom{a_5}a_8 a_9}E^{a_9}_{\phantom{a_9}a_7}(\sigma^2)^{a_5}\\
    & + 2(W^{(1)}x + b^{(1)})^{a_5}W^{(1)a_7}_{\phantom{(1)a_7}k}E^{a_5}_{\phantom{a_5}a_7}W^{(1)a_{10}}_{\phantom{(1)a_{10}}j}E^{a_5}_{\phantom{a_5}a_{10}}(\sigma^3)^{a_5})\\
    &
    W^{(2)o}_{\phantom{(2)o}a_{11}} (W^{(1)a_{11}}_{\phantom{(1)a_{11}}m}\sigma^{a_{11}} + (W^{(1)}x+b^{(1)})^{a_{11}}W^{(1)a_{12}}_{\phantom{(1)a_{12}}m}E^{a_{11}}_{\phantom{a_{11}}a_{12}}(\sigma^2)^{a_{11}}))\eta_{op}\\
    %%%%%%%%%%%%%%%%%%%%%%%%%%%%%%%%%%%%%%%%%%%%%%%%%%%%%%%%%%%%%%%%%%
    &+W^{(2)o}_{\phantom{(2)o}a_5}((W^{(1)a_5}_{\phantom{(1)a_5}m}W^{(1)a_6}_{\phantom{(1)a_6}k}E^{a_5}_{\phantom{a_5}a_6}(\sigma^2)^{a_5}+ W^{(1)a_5}_{\phantom{(1)a_5}k}W^{(1)a_7}_{\phantom{(1)a_7}m}E^{a_5}_{\phantom{a_5}a_7}(\sigma^2)^{a_5}\\
    & - (W^{(1)}x + b^{(1)})^{a_5}W^{(1)a_7}_{\phantom{(1)a_7}m}W^{(1)a_8}_{\phantom{(1)a_8}k}\delta^{a_5}_{\phantom{a_5}a_8 a_9}E^{a_9}_{\phantom{a_9}a_7}(\sigma^2)^{a_5}\\
    & + 2(W^{(1)}x + b^{(1)})^{a_5}W^{(1)a_7}_{\phantom{(1)a_7}m}E^{a_5}_{\phantom{a_5}a_7}W^{(1)a_{10}}_{\phantom{(1)a_{10}}k}E^{a_5}_{\phantom{a_5}a_{10}}(\sigma^3)^{a_5})\\
    &W^{(2)p}_{\phantom{(2)p}a_{11}} (W^{(1)a_{11}}_{\phantom{(1)a_{11}}j}\sigma^{a_{11}} + (W^{(1)}x+b^{(1)})^{a_{11}}W^{(1)a_{12}}_{\phantom{(1)a_{12}}j}E^{a_{11}}_{\phantom{a_{11}}a_{12}}(\sigma^2)^{a_{11}})\\
    & + W^{(2)p}_{\phantom{(2)p}a_5}(W^{(1)a_5}_{\phantom{(1)a_5}j}W^{(1)a_6}_{\phantom{(1)a_6}k}E^{a_5}_{\phantom{a_5}a_6}(\sigma^2)^{a_5}+ W^{(1)a_5}_{\phantom{(1)a_5}k}W^{(1)a_7}_{\phantom{(1)a_7}j}E^{a_5}_{\phantom{a_5}a_7}(\sigma^2)^{a_5}\\
    & - (W^{(1)}x + b^{(1)})^{a_5}W^{(1)a_7}_{\phantom{(1)a_7}j}W^{(1)a_8}_{\phantom{(1)a_8}k}\delta^{a_5}_{\phantom{a_5}a_8 a_9}E^{a_9}_{\phantom{a_9}a_7}(\sigma^2)^{a_5}\\
    & + 2(W^{(1)}x + b^{(1)})^{a_5}W^{(1)a_7}_{\phantom{(1)a_7}j}E^{a_5}_{\phantom{a_5}a_7}W^{(1)a_{10}}_{\phantom{(1)a_{10}}k}E^{a_5}_{\phantom{a_5}a_{10}}(\sigma^3)^{a_5})\\
    &
    W^{(2)o}_{\phantom{(2)o}a_{11}} (W^{(1)a_{11}}_{\phantom{(1)a_{11}}m}\sigma^{a_{11}} + (W^{(1)}x+b^{(1)})^{a_{11}}W^{(1)a_{12}}_{\phantom{(1)a_{12}}m}E^{a_{11}}_{\phantom{a_{11}}a_{12}}(\sigma^2)^{a_{11}}))\eta_{op}\\

    &-W^{(2)o}_{\phantom{(2)o}a_5}((W^{(1)a_5}_{\phantom{(1)a_5}k}W^{(1)a_6}_{\phantom{(1)a_6}m}E^{a_5}_{\phantom{a_5}a_6}(\sigma^2)^{a_5}+ W^{(1)a_5}_{\phantom{(1)a_5}m}W^{(1)a_7}_{\phantom{(1)a_7}k}E^{a_5}_{\phantom{a_5}a_7}(\sigma^2)^{a_5}\\
    & - (W^{(1)}x + b^{(1)})^{a_5}W^{(1)a_7}_{\phantom{(1)a_7}k}W^{(1)a_8}_{\phantom{(1)a_8}m}\delta^{a_5}_{\phantom{a_5}a_8 a_9}E^{a_9}_{\phantom{a_9}a_7}(\sigma^2)^{a_5}\\
    & + 2(W^{(1)}x + b^{(1)})^{a_5}W^{(1)a_7}_{\phantom{(1)a_7}k}E^{a_5}_{\phantom{a_5}a_7}W^{(1)a_{10}}_{\phantom{(1)a_{10}}m}E^{a_5}_{\phantom{a_5}a_{10}}(\sigma^3)^{a_5})\\
    &W^{(2)p}_{\phantom{(2)p}a_{11}} (W^{(1)a_{11}}_{\phantom{(1)a_{11}}j}\sigma^{a_{11}} + (W^{(1)}x+b^{(1)})^{a_{11}}W^{(1)a_{12}}_{\phantom{(1)a_{12}}j}E^{a_{11}}_{\phantom{a_{11}}a_{12}}(\sigma^2)^{a_{11}})\\
    & + W^{(2)p}_{\phantom{(2)p}a_5}(W^{(1)a_5}_{\phantom{(1)a_5}j}W^{(1)a_6}_{\phantom{(1)a_6}m}E^{a_5}_{\phantom{a_5}a_6}(\sigma^2)^{a_5}+ W^{(1)a_5}_{\phantom{(1)a_5}m}W^{(1)a_7}_{\phantom{(1)a_7}j}E^{a_5}_{\phantom{a_5}a_7}(\sigma^2)^{a_5}\\
    & - (W^{(1)}x + b^{(1)})^{a_5}W^{(1)a_7}_{\phantom{(1)a_7}j}W^{(1)a_8}_{\phantom{(1)a_8}m}\delta^{a_5}_{\phantom{a_5}a_8 a_9}E^{a_9}_{\phantom{a_9}a_7}(\sigma^2)^{a_5}\\
    & + 2(W^{(1)}x + b^{(1)})^{a_5}W^{(1)a_7}_{\phantom{(1)a_7}j}E^{a_5}_{\phantom{a_5}a_7}W^{(1)a_{10}}_{\phantom{(1)a_{10}}m}E^{a_5}_{\phantom{a_5}a_{10}}(\sigma^3)^{a_5})\\
    &
    W^{(2)o}_{\phantom{(2)o}a_{11}} (W^{(1)a_{11}}_{\phantom{(1)a_{11}}k}\sigma^{a_{11}} + (W^{(1)}x+b^{(1)})^{a_{11}}W^{(1)a_{12}}_{\phantom{(1)a_{12}}k}E^{a_{11}}_{\phantom{a_{11}}a_{12}}(\sigma^2)^{a_{11}}))\eta_{op}\big)
\end{array}
\end{equation}
To compute the Riemann curvature of the given manifold, one must calculate the second derivative of the metric, as required by its definition.
\begin{equation}
    R^{i}_{\phantom{i}ljk}T^{l} = [\nabla_j, \nabla_k] T^i
\end{equation}
where $\nabla_j$ is the covariant derivative, which includes the affine connection.
\begin{equation}
    \nabla_j T^i = \partial_j T^i + \Gamma^i_{\phantom{i}j l}T^l
\end{equation}
Furthermore, the bracket indicates the commutation relation between the elements; hence, the curvature can be expressed in the following way:
\begin{equation}
\begin{array}{cl}
    R^{i}_{\phantom{i}ljk}T^{l} &= \nabla_j \nabla_k T^i - \nabla_k \nabla_j T^i\\
    &= (\partial_j \Gamma^i_{\phantom{i}k l} - \partial_k \Gamma^i_{\phantom{i}j l} + \Gamma^i_{\phantom{i}jm}\Gamma^m_{\phantom{m}kl} - \Gamma^i_{\phantom{i}km}\Gamma^m_{\phantom{m}jl})T^l
\end{array}
\end{equation}
Here, the derivative of the affine connection consists of combinations of second derivatives of the metric tensor.
\begin{equation}
\begin{array}{rl} \label{dGamma}
    \partial_j \Gamma^i_{\phantom{i}k l} =& \frac{1}{2}\partial_j (g^{im}(\partial_k g_{m l} + \partial_l g_{m k} - \partial_m g_{lk}))\\
    =& \frac{1}{2}(\partial_j g^{im}(\partial_k g_{m l} + \partial_l g_{m k} - \partial_m g_{lk})\\
    & + g^{im}(\partial_j\partial_k g_{m l} + \partial_j\partial_l g_{m k} - \partial_j\partial_m g_{lk}))
\end{array}
\end{equation}
Therefore, by obtaining the specific form of the second derivative of the metric, one can express the explicit form of the Riemann curvature. To begin the computation, it is convenient to recall Eq.\ref{dg} for taking the derivative, as well as Eq.\ref{dE} and Eq.~\ref{ddE} for computing the elements involving the activation function.
\begin{equation}
\begin{array}{rl} \label{ddg}
\partial_j \partial_k g_{ml} &=
       \sum_{a_1 a_7}\big(W^{(2)o}_{\phantom{(2)o}a_1}(-2W^{(1)a_1}_{\phantom{(1)a_1}m}W^{(1)a_2}_{\phantom{(1)a_2}k}W^{(1)a_9}_{\phantom{(1)a_9}j}\delta^{a_1}_{\phantom{a_1}a_9a_{10}}E^{a_{10}}_{\phantom{a_{10}}a_2}(\sigma^2)^{a_1}\\
       & + 4W^{(1)a_1}_{\phantom{(1)a_1}m}W^{(1)a_2}_{\phantom{(1)a_2}k}W^{(1)a_9}_{\phantom{(1)a_9}j}E^{a_{1}}_{\phantom{a_{1}}a_2}E^{a_1}_{\phantom{a_{1}}a_9}(\sigma^3)^{a_1}\\
       % & - W^{(1)a_1}_{\phantom{(1)a_1}m}W^{(1)a_3}_{\phantom{(1)a_3}k}W^{(1)a_9}_{\phantom{(1)a_9}j}\delta^{a_1}_{\phantom{a_1}a_9a_{10}}E^{a_{10}}_{\phantom{a_{10}}a_3}(\sigma^2)^{a_1}\\
       % & + 2W^{(1)a_1}_{\phantom{(1)a_1}m}W^{(1)a_3}_{\phantom{(1)a_3}k}W^{(1)a_9}_{\phantom{(1)a_9}j}E^{a_{1}}_{\phantom{a_{1}}a_3}E^{a_1}_{\phantom{a_{1}}a_9}(\sigma^3)^{a_1}\\
       & -W^{(1)a_1}_{\phantom{(1)a_1}j}W^{(1)a_3}_{\phantom{(1)a_3}m}W^{(1)a_4}_{\phantom{(1)a_4}k}\delta^{a_1}_{\phantom{a_1}a_4a_5}E^{a_5}_{\phantom{a_5}a_3}(\sigma^2)^{a_1}\\
       & + (W^{(1)}x + b^{(1)})^{a_1}W^{(1)a_3}_{\phantom{(1)a_3}m}W^{(1)a_4}_{\phantom{(1)a_4}k}W^{(1)a_9}_{\phantom{(1)a_9}j}\delta^{a_1}_{\phantom{a_1}a_4a_5}\delta^{a_5}_{\phantom{a_5}a_9a_{10}}E^{a_{10}}_{\phantom{a_{10}}a_3}(\sigma^2)^{a_1}\\
       & - 2(W^{(1)}x + b^{(1)})^{a_1}W^{(1)a_3}_{\phantom{(1)a_3}m}W^{(1)a_4}_{\phantom{(1)a_4}k}W^{(1)a_9}_{\phantom{(1)a_9}j}\delta^{a_1}_{\phantom{a_1}a_4a_5}E^{a_5}_{\phantom{a_5}a_3}E^{a_1}_{\phantom{a_1}a_9}(\sigma^3)^{a_1}\\
       & + 2W^{(1)a_1}_{\phantom{(1)a_1}j}W^{(1)a_3}_{\phantom{(1)a_3}m}W^{(1)a_6}_{\phantom{(1)a_6}k}E^{a_1}_{\phantom{a_1}a_3}E^{a_1}_{\phantom{a_1}a_6}(\sigma^3)^{a_1}\\
       % & - 2(W^{(1)}x + b^{(1)})^{a_1}W^{(1)a_3}_{\phantom{(1)a_3}m}W^{(1)a_9}_{\phantom{(1)a_9}j}W^{(1)a_6}_{\phantom{(1)a_6}k} \delta^{a_1}_{\phantom{a_1}a_9 a_{10}} E^{a_{10}}_{\phantom{a_{10}}a_3}  E^{a_{1}}_{\phantom{a_{1}}a_6} (\sigma^3)^{a_1}\\
       & - 4(W^{(1)}x + b^{(1)})^{a_1}W^{(1)a_3}_{\phantom{(1)a_3}m} W^{(1)a_6}_{\phantom{(1)a_6}k} W^{(1)a_9}_{\phantom{(1)a_9}j} \delta^{a_1}_{\phantom{a_1}a_9 a_{10}} E^{a_{1}}_{\phantom{a_{1}}(a_3} E^{a_{10}}_{\phantom{a_{10}}a_6)} (\sigma^3)^{a_1}\\
       & + 6(W^{(1)}x + b^{(1)})^{a_1}W^{(1)a_3}_{\phantom{(1)a_3}m}W^{(1)a_6}_{\phantom{(1)a_6}k}W^{(1)a_9}_{\phantom{(1)a_9}j}E^{a_1}_{\phantom{a_1}a_9}E^{a_1}_{\phantom{a_1}a_3}E^{a_1}_{\phantom{a_1}a_6}(\sigma^4)^{a_1} )\\
       & W^{(2)p}_{\phantom{(2)p}a_7} (W^{(1)a_7}_{\phantom{(1)a_7}l}\sigma^{a_7} + (W^{(1)}x+b^{(1)})^{a_7}W^{(1)a_8}_{\phantom{(1)a_8}l}E^{a_7}_{\phantom{a_7}a_8}(\sigma^2)^{a_7})\\
       & + W^{(2)o}_{\phantom{(2)o}a_1}(W^{(1)a_1}_{\phantom{(1)a_1}m}W^{(1)a_2}_{\phantom{(1)a_2}k}E^{a_1}_{\phantom{a_1}a_2}(\sigma^2)^{a_1}+ W^{(1)a_1}_{\phantom{(1)a_1}k}W^{(1)a_3}_{\phantom{(1)a_3}m}E^{a_1}_{\phantom{a_1}a_3}(\sigma^2)^{a_1}\\
    & - (W^{(1)}x + b^{(1)})^{a_1}W^{(1)a_3}_{\phantom{(1)a_3}m}W^{(1)a_4}_{\phantom{(1)a_4}k}\delta^{a_1}_{\phantom{a_1}a_4 a_5}E^{a_5}_{\phantom{a_5}a_3}(\sigma^2)^{a_1}\\
    & + 2(W^{(1)}x + b^{(1)})^{a_1}W^{(1)a_3}_{\phantom{(1)a_3}m}W^{(1)a_6}_{\phantom{(1)a_6}k}E^{a_1}_{\phantom{a_1}a_3}E^{a_1}_{\phantom{a_1}a_6}(\sigma^3)^{a_1})\\
      & W^{(2)p}_{\phantom{(2)p}a_7}(W^{(1)a_7}_{\phantom{(1)a_7}l} W^{(1)a_7}_{\phantom{(1)a_7}a_{11}}E^{a_{11}}_{\phantom{a_{11}}j}(\sigma^2)^{a_7} + W^{(1)a_7}_{\phantom{(1)a_7}j}W^{(1)a_8}_{\phantom{(1)a_8}l}E^{a_7}_{\phantom{a_7}a_8}(\sigma^2)^{a_7}\\
       & - (W^{(1)}x + b^{(1)})^{a_7}W^{(1)a_8}_{\phantom{(1)a_8}l}W^{(1)a_{11}}_{\phantom{(1)a_{11}}j}\delta^{a_{7}}_{\phantom{a_{7}}a_{11}a_{12}}E^{a_{12}}_{\phantom{a_{12}}a_8}(\sigma^2)^{a_7}\\
       & + 2(W^{(1)}x + b^{(1)})^{a_7}W^{(1)a_8}_{\phantom{(1)a_8}l}W^{(1)a_7}_{\phantom{(1)a_7}a_{11}}E^{a_7}_{\phantom{a_7}a_8}E^{a_{11}}_{\phantom{a_{11}}j}(\sigma^3)^{a_7})\\
       & + W^{(2)p}_{\phantom{(2)p}a_1}(- 2W^{(1)a_1}_{\phantom{(1)a_1}(l} W^{(1)a_2}_{\phantom{(1)a_2}k)}W^{(1)a_{11}}_{\phantom{(1)a_{11}}j}\delta^{a_{1}}_{\phantom{a_{1}}a_{11}a_{12}}E^{a_{12}}_{\phantom{a_{12}}a_2}(\sigma^2)^{a_1}\\
       &+ 4 W^{(1)a_1}_{\phantom{(1)a_1}(l} W^{(1)a_2}_{\phantom{(1)a_2}k)} W^{(1)a_{11}}_{\phantom{(1)a_{11}}j} E^{a_1}_{\phantom{a_1}a_2} E^{a_1}_{\phantom{a_1}a_{11}} (\sigma^3)^{a_1} \\
       % &- W^{(1)a_1}_{\phantom{(1)a_1}k} W^{(1)a_3}_{\phantom{(1)a_3}l} W^{(1)a_{11}}_{\phantom{(1)a_{11}}j} \delta^{a_{1}}_{\phantom{a_{1}}a_{11}a_{12}} E^{a_{12}}_{\phantom{a_{12}}a_3} (\sigma^2)^{a_1}\\
       % &+ 2 W^{(1)a_1}_{\phantom{(1)a_1}k} W^{(1)a_3}_{\phantom{(1)a_3}l} W^{(1)a_{11}}_{\phantom{(1)a_{11}}j} E^{a_1}_{\phantom{a_1}a_3} E^{a_1}_{\phantom{a_1}a_{11}} (\sigma^3)^{a_1} \\
       & - W^{(1)a_1}_{\phantom{(1)a_1}j} W^{(1)a_3}_{\phantom{(1)a_3}l} W^{(1)a_4}_{\phantom{(1)a_4}k} \delta^{a_1}_{\phantom{a_1}a_4a_5} E^{a_5}_{\phantom{a_5}a_3} (\sigma^2)^{a_1}\\
       & + (W^{(1)}x + b^{(1)})^{a_1} W^{(1)a_3}_{\phantom{(1)a_3}l} W^{(1)a_4}_{\phantom{(1)a_4}k} W^{(1)a_{11}}_{\phantom{(1)a_{11}}j} \delta^{a_1}_{\phantom{a_1}a_4a_5}  \delta^{a_{5}}_{\phantom{a_{5}}a_{11} a_{12}} E^{a_{12}}_{a_3} (\sigma^2)^{a_1} \\
       & - 2 (W^{(1)}x + b^{(1)})^{a_1} W^{(1)a_3}_{\phantom{(1)a_3}l} W^{(1)a_{11}}_{\phantom{(1)a_{11}}j} W^{(1)a_4}_{\phantom{(1)a_4}k} \delta^{a_1}_{\phantom{a_1}a_4a_5} E^{a_5}_{\phantom{a_5}a_3} E^{a_1}_{\phantom{a_1}a_{11}} (\sigma^3)^{a_1}\\
       & + 2 W^{(1)a_1}_{\phantom{(1)a_1}j} W^{(1)a_3}_{\phantom{(1)a_3}l} W^{(1)a_6}_{\phantom{(1)a_6}k} E^{a_1}_{\phantom{a_1}a_3} E^{a_1}_{\phantom{a_1}a_6} (\sigma^3)^{a_1} \\
       & - 4 (W^{(1)}x + b^{(1)})^{a_1} W^{(1)a_3}_{\phantom{(1)a_3}l} W^{(1)a_6}_{\phantom{(1)a_6}k} W^{(1)a_{11}}_{\phantom{(1)a_{11}}j}  \delta^{a_{1}}_{\phantom{a_{1}}a_{11}a_{12}} E^{a_{12}}_{\phantom{a_{12}}(a_3} E^{a_1}_{\phantom{a_1}a_6)} (\sigma^3)^{a_1}\\
       % & - 2 (W^{(1)}x + b^{(1)})^{a_1} W^{(1)a_3}_{\phantom{(1)a_3}l} W^{(1)a_6}_{\phantom{(1)a_6}k} W^{(1)a_1}_{\phantom{(1)a_1}a_{11}} \delta^{a_{11}}_{\phantom{a_{11}}ja_{12}} E^{a_1}_{\phantom{a_1}a_3} E^{a_{12}}_{\phantom{a_{12}}a_6} (\sigma^3)^{a_1} \\
       & + 6 (W^{(1)}x + b^{(1)})^{a_1} W^{(1)a_3}_{\phantom{(1)a_3}l} W^{(1)a_6}_{\phantom{(1)a_6}k} W^{(1)a_{11}}_{\phantom{(1)a_{11}}j} E^{a_1}_{\phantom{a_1}a_3} E^{a_1}_{\phantom{a_1}a_6} E^{a_1}_{\phantom{a_1}a_{11}} (\sigma^4)^{a_1})\\
       & W^{(2)o}_{\phantom{(2)o}a_7} (W^{(1)a_7}_{\phantom{(1)a_7}m}\sigma^{a_7} + (W^{(1)}x+b^{(1)})^{a_7}W^{(1)a_8}_{\phantom{(1)a_8}m}E^{a_7}_{\phantom{a_7}a_8}(\sigma^2)^{a_7})\\
       & + W^{(2)p}_{\phantom{(2)p}a_1}(W^{(1)a_1}_{\phantom{(1)a_1}l}W^{(1)a_2}_{\phantom{(1)a_2}k}E^{a_1}_{\phantom{a_1}a_2}(\sigma^2)^{a_1}+ W^{(1)a_1}_{\phantom{(1)a_1}k}W^{(1)a_3}_{\phantom{(1)a_3}l}E^{a_1}_{\phantom{a_1}a_3}(\sigma^2)^{a_1}\\
    & - (W^{(1)}x + b^{(1)})^{a_1}W^{(1)a_3}_{\phantom{(1)a_3}l}W^{(1)a_4}_{\phantom{(1)a_4}k}\delta^{a_1}_{\phantom{a_1}a_4 a_5}E^{a_5}_{\phantom{a_5}a_3}(\sigma^2)^{a_1}\\
    & + 2(W^{(1)}x + b^{(1)})^{a_1}W^{(1)a_3}_{\phantom{(1)a_3}l}W^{(1)a_6}_{\phantom{(1)a_6}k} E^{a_1}_{\phantom{a_1}a_3} E^{a_1}_{\phantom{a_1}a_6}(\sigma^3)^{a_1})\\
    &W^{(2)o}_{\phantom{(2)o}a_7}(2W^{(1)a_7}_{\phantom{(1)a_7}(m} W^{(1)a_{11}}_{\phantom{(1)a_{11}}j)}E^{a_7}_{\phantom{a_7}a_{11}}(\sigma^2)^{a_7} 
    % + W^{(1)a_7}_{\phantom{(1)a_7}j} W^{(1)a_8}_{\phantom{(1)a_8}m} E^{a_7}_{\phantom{a_7}a_8} (\sigma^2)^{a_7}
    \\
    & - (W^{(1)}x + b^{(1)})^{a_7} W^{(1)a_8}_{\phantom{(1)a_8}m} W^{(1)a_{11}}_{\phantom{(1)a_{11}}j} \delta^{a_{7}}_{\phantom{a_{7}}a_{11}a_{12}} E^{a_{12}}_{\phantom{a_{12}}a_8} (\sigma^2)^{a_7}\\
       & + 2 (W^{(1)} x + b^{(1)})^{a_7} W^{(1)a_8}_{\phantom{(1)a_8}m} W^{(1)a_{11}}_{\phantom{(1)a_{11}}j} E^{a_7}_{\phantom{a_7}a_8} (\sigma^3)^{a_7}  E^{a_7}_{\phantom{a_7}a_{11}} \big)\eta_{op}
    
\end{array}
\end{equation}
Now, we have collected all the fundamental components needed to compute the Ricci scalar. Although the computation is quite tedious, it can be carried out through brute-force calculation by referring to Eqs.~\ref{Riemman}, \ref{curvatures}, \ref{dg}, \ref{inv_metric}, \ref{Gamma}, \ref{dGamma}, and \ref{ddg}.

%%%%%%%%%%%%%%%%%%%%%%%%%%%%%%%%%%%%%%%%%%%%%%%%%%%%%%%%%%%%%%%%%%%%%%%%%%%%%%%%%%%%%%%%%%%%%%%%%%%%%%%%%%%%%

\subsection{Quadratic Case for Computation Check}
This entire sequence is indeed both tedious and complex. Therefore, we introduce the simplest case for each process to verify the validity of the code and the formulas. Here, we set the activation function to the quadratic of the input signal and maintain the number of layers at two. Under these conditions, the transformed vector becomes:
\begin{equation}
    x'^i = W^{(2)i}_{\phantom{(2)i}j}(W^{(1)} x + b^{(1)})^{2j} + b^{(2)i}
\end{equation}
Now, the Jacobian can be easily derived from the above equation.
\begin{equation} \label{jaco3}
    J^{i}_{\phantom{i}j} = W^{(2)i}_{\phantom{(2)i}k} 2(W^{(1)} x + b^{(1)})^k_{\phantom{k}m} W^{(1)m}_{\phantom{(1)m}j}
\end{equation}
Here, $(W^{(1)}x + b^{(1)})^i_{\phantom{i}j}$ has a diagonal matrix form as follows:
\begin{equation}
    (W^{(1)}x + b^{(1)})^i_{\phantom{i}j} = \Bigg\{
\begin{array}{cl}
     W^{(1)i}_{\phantom{(1)i}k} x^k + b^{(1)i} &  \textrm{if $i = j$}\\
     0 & \textrm{if $ i \neq j$}
\end{array}
\end{equation}
Then, the metric can be written in the following form:
\begin{equation}
    g_{ij} = W^{(2)l}_{\phantom{(2)l}k}2(W^{(1)} x + b^{(1)})^k_{\phantom{k}m}W^{(1)m}_{\phantom{(1)m}i}
    W^{(2)}_{\phantom{(2)}ln}2(W^{(1)} x + b^{(1)})^n_{\phantom{n}o}W^{(1)o}_{\phantom{(1)o}j}
\end{equation}
Finally, one can compute the derivative of the metric.
\begin{equation}
\begin{array}{ll}
    \partial_k g_{ij} =& 4 |W^{(2)o}_{\phantom{(2)o}q}|^2(W^{(1)})^{2q}_{\phantom{2q}p} \delta^p_{\phantom{p}ik}
    W^{(1)n}_{\phantom{(1)i}o}(W^{(1)} x + b^{(1)})^o_{\phantom{o}j}\\
    &+ 4 |W^{(2)}_{\phantom{(2)}nq}|^2W^{(1)q}_{\phantom{(1)i}m}(W^{(1)} x + b^{(1)})^m_{\phantom{m}i}(W^{(1)})^{2n}_{\phantom{2n}p}\delta^p_{\phantom{p}jk}
\end{array}
\end{equation}
where $|W|^2 = W^T W$ and $W^2 = W^i_{\phantom{i}k} W^k_{\phantom{k}j}$.
\subsubsection{2-Dim Simplest Example for Square Activation}
To cross-check the computation results, we hereby introduce the simplest example for metric computation in 2D. The weights and biases for each layer are defined as follows:
\begin{eqnarray}
    W^{(1)i}_{\phantom{(1)i}j} = \left(\begin{array}{cc}
        1 & 2 \\
        3 & 4
    \end{array}\right)\\
    W^{(2)i}_{\phantom{(1)i}j} = \left(\begin{array}{cc}
        5 & 6 \\
        7 & 8
    \end{array}\right)\\
    b^{(1)i} = \left(\begin{array}{c}
        3 \\
        4
    \end{array}\right)\\
    x = \left(\begin{array}{c}
        1 \\
        2
    \end{array}\right)
\end{eqnarray}
Then, the Jacobian can be computed as expressed in Eq.~\ref{jaco3}. We will break down the Jacobian piece by piece and verify the validity of the equation.
\begin{equation}
    2(W^{(1)} x + b^{(1)})^m_{\phantom{m}j} = \left(\begin{array}{cc}
        16 & 0 \\
        0 & 30
    \end{array}\right)
\end{equation}
By multiplying the weights from both layers, the equation becomes:
\begin{equation}
    J^i_{\phantom{i}j} = \left(\begin{array}{cc}
        620 & 880 \\
        832 & 1184
    \end{array}\right)
\end{equation}
Finally, the metric can be expressed as the square of the Jacobian.
\begin{equation}
    g_{ij} = (J^{T})_{ik} J^k_{\phantom{k}j} = \left(\begin{array}{cc}
        1076624 & 1530688 \\
        1530688 & 2176256
    \end{array}\right)
\end{equation}
As shown above, the metric is symmetric.

\subsubsection{2-Dim Simplest Example for SiLU Activation}
In practice, a simple square activation is insufficient to capture the complex structure of curved space. Therefore, we adopt the SiLU activation function to better express the model's geometric structure. The SiLU function behaves similarly to the ReLU activation but enjoys smoothness across the entire domain.
\begin{equation}
    \textrm{SiLU}(x) = x \times \textrm{LS}(x) = \frac{x}{1+e^{-x}}
\end{equation}
Since the SiLU activation contains the inverse of the exponential function in its expression, the input values should be kept smaller than 1 to prevent the activation from converging to a trivial value.
\begin{eqnarray}
    W^{(1)i}_{\phantom{(1)i}j} = \left(\begin{array}{cc}
        0.1 & 0.2 \\
        0.3 & 0.4
    \end{array}\right)\\
    W^{(2)i}_{\phantom{(1)i}j} = \left(\begin{array}{cc}
        0.5 & 0.6 \\
        0.7 & 0.8
    \end{array}\right)\\
    b^{(1)i} = \left(\begin{array}{c}
        0.3 \\
        0.4
    \end{array}\right)\\
    x = \left(\begin{array}{c}
        0.1 \\
        0.2
    \end{array}\right)
\end{eqnarray}
Using the input example set described above, one can compute the explicit equations, with the results as follows. We will first introduce the main components used in the calculation. One key component is the sigmoid function, which is utilized in the SiLU computation.
\begin{equation}
    \sigma^i = \left(\begin{array}{c}
        0.5866 \\
        0.6248
    \end{array}\right)
\end{equation}
Another component is the diagonalized exponential term, which appears in the derivative of the vector exponential.
\begin{equation}
    E^{i}_{\phantom{i}j} = \left(\begin{array}{cc}
        0.7047 & 0\\
        0 & 0.6005
    \end{array}\right)
\end{equation}
By combining the two expressions above with the weights and biases, it is possible to obtain the full Jacobian.
\begin{equation}
    J^i_{\phantom{i}j} = \left(\begin{array}{cc}
        0.1676 & 0.2458\\
        0.2257 & 0.3322
    \end{array}\right)
\end{equation}
Finally, by squaring the Jacobian, the induced metric $g_{ij}$ can be defined.
\begin{equation}
    g_{ij} = (J^{T})_{ik} J^k_{\phantom{k}j} = \left(\begin{array}{cc}
        0.0790 & 0.1161 \\
        0.1161 & 0.1708
    \end{array}\right)
\end{equation}
As shown above, the metric is well-defined and forms a symmetric structure in this setup as well.

%%%%%%%%%%%%%%%%%%%%%%%%%%%%%%%%%%%%%%%%%%%%%%%%%%%%%%%%%%%%%%%%%%%%%%%%%%%%%%%%%%%%%
%%%%%%%%%%%%%%%%%%%%%%%%%%%%%%%%%%%%%%%%%%%%%%%%%%%%%%%%%%%%%%%%%%%%%%%%%%%%%%%%%%%%%
%%%%%%%%%%%%%%%%%%%%%%%%%%%%%%%%%%%%%%%%%%%%%%%%%%%%%%%%%%%%%%%%%%%%%%%%%%%%%%%%%%%%%

\section{Base Graph Neural Network Model}
\label{gnn}
In general, molecule is represented in a graph form. Therefore, in order to handle molecule dataset, it is inevitable to utilize graph neural networks. We chose directional message passing network (DMPNN) \citet{dmpnn} for our backbone, since it outperforms other GNN architectures in molecular domain. Given a graph, DMPNN initializes the hidden state of each edge $(i, j)$ based on its edge feature $E_{ij}$ with node feature $X_i$. At each step $t$, directional edge summarizes incident edges as a message $m_{ij}^{t+1}$ and updates its hidden state to $h_{ij}^{t+1}$.
\begin{gather}
    m_{ij}^{t+1} = \sum_{k \in {\mathcal{N}}(i) \backslash j}h_{ki}^{t} \\
    h_{ij}^{t+1} = \mathrm{ReLU}(h_{ij}^{0} + W_em_{ij}^{t+1})
\end{gather}
Where $\mathcal{N}(i)$ denotes the set of neighboring nodes and $W_e$ a learnable weight.he hidden states of nodes are updated by aggregating the hidden states of incident edges into message $m_i^{t+1}$, and passing its concatenation with the node feature $X_i$ into a linear layer followed by ReLU non-linearity
\begin{gather}
    m_i^{t+1} = \sum_{j \in \mathcal{N}(i)} h_{ij}^t\\
    h_i^{t+1} = \mathrm{ReLU}(W_n \mathrm{concat}(X_i, m_i^{t+1})) 
\end{gather}
Similarly, $W_n$ denotes a learnable weight. Assuming DMPNN runs for $T$ timesteps, we use $(X_{out},E_{out}) = \mathrm{GNN}(A, X, E)$ to denote the output representation matrices containing hidden states of all nodes and edges, respectively (i.e., $X_{out,i} = h_i^T$ and $E_{out,ij} = h_{ij}^T$).

For graph-level prediction, the node representations after the final GNN layer are typically sum-pooled to obtain a single graph representation $h_G = \sum_{i} h_i$, which is then passed to a FFN prediction layer. 

% This forms diffeomorphism groups generated by Lie derivative.
% \begin{equation}
% \begin{array}{ll}
%         (\mathcal{L}_XT)^{a_1 \cdots a_r}_{\phantom{a_1 \cdots a_r} b_1 \cdots b_s} =& \sum_c[X^c(\partial_c T^{a_1 \cdots a_r}_{\phantom{a_1 \cdots a_r} b_1 \cdots b_s})
%     \\
%     & - (\partial_cX^{a_1})T^{ca_2 \cdots a_r}_{\phantom{ca_2 \cdots a_r} b_1 \cdots b_s} - \cdots - (\partial_cX^{a_r})T^{a_1 \cdots c}_{\phantom{a_1 \cdots a_r} b_1 \cdots b_s}
%     \\
%     &  + (\partial_{b_1}X^c)T^{a_1 \cdots a_r}_{\phantom{a_1 \cdots a_r} cb_2 \cdots b_s} + \cdots + (\partial_{b_s}X^c)T^{a_1 \cdots a_r}_{\phantom{a_1 \cdots a_r} b_1 \cdots c}]
% \end{array}
% \end{equation}

%%%%%%%%%%%%%%%%%%%%%%%%%%%%%%%%%%%%%%%%%%%%%%%%%%%%%%%%%%%%%%%%%%%%%%%%%%%%
%%%%%%%%%%%%%%%%%%%%%%%%%%%%%%%%%%%%%%%%%%%%%%%%%%%%%%%%%%%%%%%%%%%%%%%%%%%%
%%%%%%%%%%%%%%%%%%%%%%%%%%%%%%%%%%%%%%%%%%%%%%%%%%%%%%%%%%%%%%%%%%%%%%%%%%%%
% 논문에서 표기한 네트워크 이름 / 실제 구현된 구조
% embedding 이 backbone front depth2
% backbone 을 없애고 
% encoder 는 backbone back + botteneck(의 encoder)

\section{Architecture and Hyperparameters}
\label{architecture}
\begin{figure}[ht]
\begin{center}
\includegraphics[trim={0cm 4cm 0cm 4cm}, clip, width=1\linewidth]{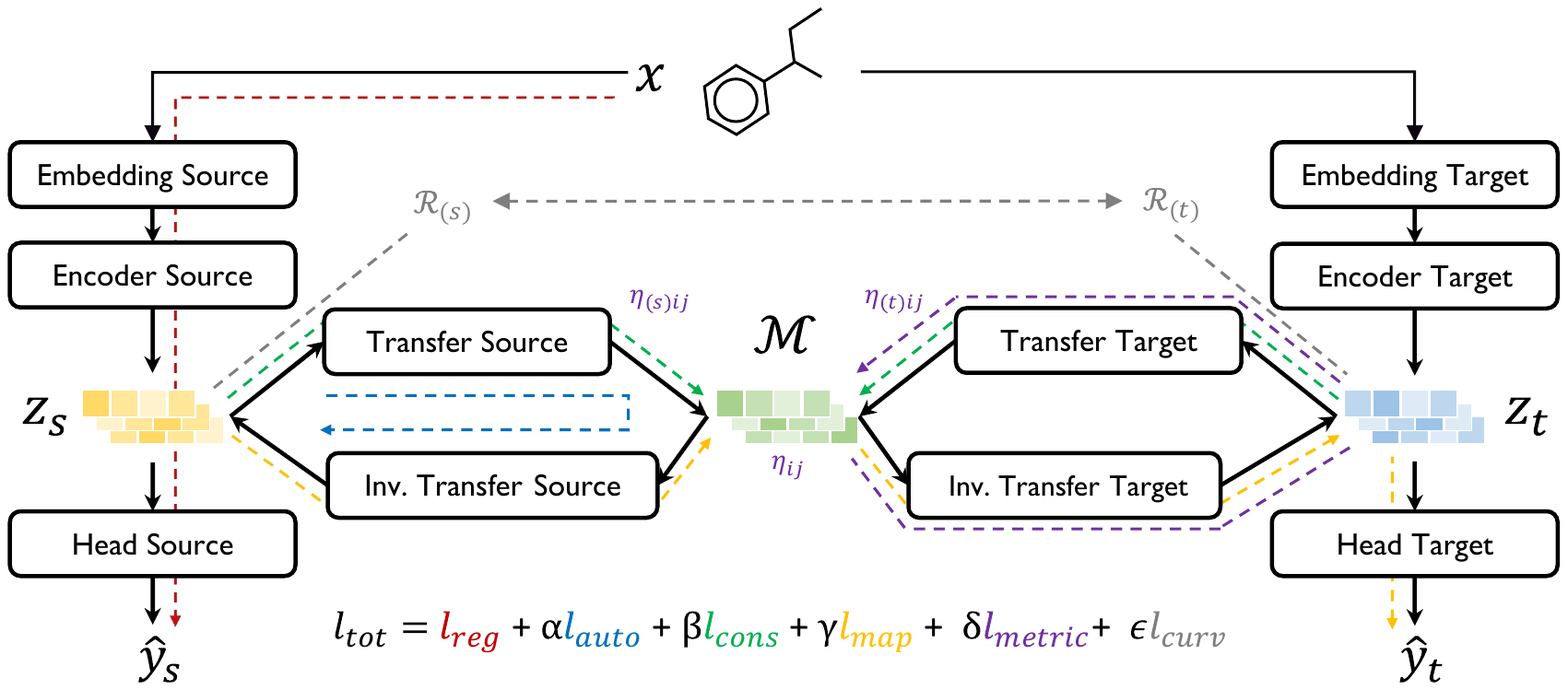}
\end{center}
\caption{Detailed schematics of GEAR with specific loss function components.}
\label{fig:apdx_algo}
\end{figure}

% Detailed steps of training {\gate} is described in Algorithm ~\ref{algorithm}. The architecture of our model is composed of five distinct networks and their parameter sizes are depicted in Table \ref{network}. As illustrated in Figure~\ref{fig:fig2}, one backbone network is shared across tasks, and the bottleneck, transfer, inverse transfer, and head network exists for each task. The backbone network $embedd(\cdot)$ has the DMPNN architecture with depth 4, and it converts the input molecule representation $x$ into a new representation $a$ in a common embedding space. We apply perturbation $perturb(\cdot)$ to $a$ for a number of perturbations, which is set to 10 in this paper. All of the perturbed representation $\left\{{\bar{a}}\right\}$ along with $a$ are then fed into the bottleneck network. The bottleneck network has an autoencoder structure and the output from the encoder $f_e(a)$ becomes the input to the transfer network and head network. %Transfer, inverse transfer, and head network are composed of MLP layers.
Detailed steps of training GEAR are described in Algorithm ~\ref{algorithm}. The architecture of our model is composed of five distinct networks and their parameter sizes are depicted in Table \ref{network}. As illustrated in Figures~\ref{fig:algo_figure} and \ref{fig:apdx_algo}, each task is equipped with five modules: an embedding network, encoder network, transfer network, inverse transfer network, and head network. The embedding network $embedd(\cdot)$ has the DMPNN architecture with task-specific model structure and converts the input molecule representation $x$ into a new representation $a$ in a embedding space. The encoder network is composed of a bottleneck network. The bottleneck network has an autoencoder structure with MLP layers. The output from the encoder $f_e(a)$ becomes the input to the transfer network and head network. %Transfer, inverse transfer, and head network are composed of MLP layers. 
The output of transfer network $f_t(z)$, denoted as $m$, is used to calculate consistency loss. It is also fed into inverse transfer network, so that the output from inverse transfer network $f_i(m)$ can be used to calculate autoencoder loss. Both modules are utilized to compute mapping, metric and curvature losses. The output from head network, ${f_{h}}\circ {f_{i}(m)}$, is used to calculate regression loss and mapping loss. We trained the model for 1000 epochs with batch size 512 while using AdamW \citet{loshchilov2017decoupled} for optimization with learning rate 5e-5. %ReduceLROnPlateau with lower bound of learning rate 1e-5, patience 20, and threshold 0.001 was used for scheduling. 
The hyperparameters for $\alpha, \beta, \gamma, \delta, \epsilon$ are 0.1, 0.1, 0.2, 0.1, 0.2 respectively.

\begin{algorithm}[t!]
	\caption{GEAR}
        \label{algorithm}
	\begin{algorithmic}[1]
        \State Initialize embedding network ${f}_{m}$, encoder network ${f}_{e}$, transfer network ${f}_{t}$, inverse transfer network ${f}_{i}$, head network ${f}_{h}$ with random parameters
        $\theta$
        \State Let $\mathcal{J}(\cdot)$, $\mathcal{G}(\cdot)$, and $\mathcal{R}(\cdot)$ be mathematical functions for jacobian, metric, and curvature computation
        \\
        \For {epoch $i=1,2,\ldots n$}
            \For {each $t \in Tasks$}
            \State Initialize $L_{metric}, L_{reg}, L_{auto}$ to 0
                \For {each batch $\mathbf{b}= (x^{t},y^{t}) \in  $ dataset $D $}
                \State $a^{t} \leftarrow {f}^{m}_{e}(x^{t})$
                % \State $\left\{{\bar{a}}^{t}\right\} \leftarrow perturb(a^{t})$\\
                \State $z^{t} \leftarrow {f}^{t}_{e}(a^{t})$
                \State $m^{t} \leftarrow {f}^{t}_{t}(z^{t})$
                \State $g^{t}_{curved} \leftarrow \mathcal{G}(\mathcal{J}(m^t, f_i^t))$
                \State $g^{t}_{flat} \leftarrow \mathcal{G}(\mathcal{J}({f}^{t}_{i}(m^{t}, f_t^t), g^{t}_{curved})$
                \State $r^t \leftarrow \mathcal{R}(g^{t}_{curved})$
                \For{step $k = 1,\dots,K$}
                    \State $g^{t,\,(k)}_{\text{curved}} \leftarrow \mathcal{G}(\mathcal{J}(m^t, f_i^t), g^{t,\,(k-1)}_{\text{flat}})$
                    \State $g^{t,\,(k)}_{\text{flat}} \leftarrow \mathcal{G}(\mathcal{J}(f^{t}_{i}(m^{t}, f^{t}_{t})), g^{t,\,(k-1)}_{\text{curved}})$
                \EndFor
                \State $L_{metric} \leftarrow \sum_{k=1}^{K} \left( MSELoss(g^{t,\,(k)}_{\text{flat}},\ I) + MSELoss(g^{t,\,(k)}_{\text{curved}},\ I) \right)$
                \State $L_{reg} \leftarrow MSE Loss(y^{t},{f}^{t}_{h}(z^{t})) $
                \State $L_{auto} \leftarrow MSE Loss({f}^{t}_{i}(m^{t}),z^{t}) $\\

                \For {each $s \in Subtasks$}
                \State Initialize $L_{map}, L_{cons}, L_{curv}$ to 0
                \State $z^{s} \leftarrow {f}^{s}_{e}(a^{t})$
                \State $m^{s} \leftarrow {f}^{s}_{t}(z^{s})$
                \State $g^{s}_{curved} \leftarrow \mathcal{G}(\mathcal{J}(m^s, f_i^s))$
                \State $g^{s}_{flat} \leftarrow \mathcal{G}(\mathcal{J}({f}^{s}_{i}(m^{s}, f_s^t), g^{s}_{curved})$
                \State $r^s \leftarrow \mathcal{R}(g^{s}_{curved})$
                \For{step $k = 1,\dots,K$}
                    \State $g^{s,\,(k)}_{\text{curved}} \leftarrow \mathcal{G}(\mathcal{J}(m^s, f_i^s), g^{s,\,(k-1)}_{\text{flat}})$
                    \State $g^{s,\,(k)}_{\text{flat}} \leftarrow \mathcal{G}(\mathcal{J}(f^{s}_{i}(m^{s}, f^{s}_{t})), g^{s,\,(k-1)}_{\text{curved}})$
                \EndFor
                \State $L_{metric} \leftarrow L_{metric} + \sum_{k=1}^{K} \left( MSELoss(g^{s,\,(k)}_{\text{flat}},\ I) + MSELoss(g^{s,\,(k)}_{\text{curved}},\ I) \right)$
                \State $L_{map} \leftarrow L_{map} + MSE Loss(y^{t},{f}^{t}_{h}\circ{f}^{t}_{i}(m^{s}))$
                \State $L_{cons} \leftarrow L_{cons} + MSE Loss(m^{t},m^{s})$
                \State $L_{curv} \leftarrow L_{curv} + MSELoss(r^t, r^s)$
                \EndFor\\
                % \State Compute $L_{total} = L_{reg} + \alpha L_{auto} + \beta L_{map} + \gamma L_{cons} + \delta L_{metric} +\epsilon L_{curv}$
\State \textbf{Compute} \text{$L_{{total}} = L_{{reg}} + \alpha L_{{auto}} + \beta L_{{map}} + \gamma L_{{cons}} + \delta L_{{metric}} + \epsilon L_{{curv}}$}

                \State Update $\theta$ using $L_{total}$
                
            \EndFor
        \EndFor
    \EndFor
	\end{algorithmic} 
\end{algorithm}

\begin{table}[!ht]
\caption{Common Network Parameters}
\label{network}
\begin{center}
    \begin{tabular}{ccccc}
    \hline
        \textbf{network} & \textbf{layer} & \textbf{input, output size} & \textbf{hidden size} & \textbf{dropout } \\ \hline
        bottleneck & MLP layer & 100, 50 & 50 & 0  \\ \hline
        transfer & MLP layer & 50, 50 & 50,50,50 & 0.2  \\ \hline
        inverse transfer & MLP layer & 50, 50 & 50,50,50 & 0.2  \\ \hline
        head & MLP layer & 50, 1 & 25,12 & 0.2  \\ \hline
    \end{tabular}
\end{center}
\end{table}

\begin{table}[!ht]
    \caption{Task Specific Encoder Parameters}
    \label{task_specific_param}
    \begin{center}
    \begin{tabular}{ c  c c}
       Tasks  & Random Split Encoder Parameters & Scaffold Split Encoder Parameters\\
         \hline
\bf{hv $\leftarrow$ ds} & [200, 200] & [200, 200, 200]\\
\bf{as $\leftarrow$ bp} & [200, 200]& [200, 200, 200]\\
\bf{ds $\leftarrow$ kri} & [200]& [200, 200]\\
\bf{hv $\leftarrow$ vs} & [200, 200]& [200]\\
\bf{vs $\leftarrow$ hv} & [200, 200, 200]& [200]\\
\bf{st $\leftarrow$ as} & [200, 200, 200]& [200, 200, 200]\\
\bf{ds $\leftarrow$ lp} & [200, 200, 200]& [200]\\
\bf{pol $\leftarrow$ ds} & [200, 200]& [200, 200, 200]\\
\bf{vs $\leftarrow$ bp} & [200, 200, 200]& [200, 200, 200]\\
\bf{dk $\leftarrow$ ef} & [200]& [200, 200]\\
\bf{as $\leftarrow$ ccs} & [200]& [200, 200, 200]\\
\bf{ct $\leftarrow$ bp} & [200, 200]& [200, 200]\\
\bf{st $\leftarrow$ ccs} & [200, 200, 200]& [200]\\
\bf{ccs $\leftarrow$ kri} & [200]& [200, 200, 200]\\
\bf{hv $\leftarrow$ bp} & [200, 200, 200]& [200, 200, 200]\\
\bf{vs $\leftarrow$ ccs} & [200, 200, 200]& [200]\\
\bf{st $\leftarrow$ hv} & [200, 200]& [200]\\
\bf{hv $\leftarrow$ ct} & [200, 200]& [200, 200]\\
\bf{ip $\leftarrow$ bp} & [200]& [200]\\
\bf{hv $\leftarrow$ ef} & [200, 200]& [200, 200]\\
\bf{hv $\leftarrow$ kri} & [100, 100, 100]& [200, 200]\\
\bf{ct $\leftarrow$ kri} & [200, 200]& [200, 200, 200]\\
\bf{ip $\leftarrow$ dk} &  [200]& [200, 200, 200]\\
\hline

\end{tabular}
\end{center}
\end{table}

\begin{table}[ht]
\caption{Hyperparameters}
\label{hyperparameter}
\begin{center}
    \begin{tabular}{cc}
    \hline
        % \multicolumn{2}{|c|}{\textbf{Hyperparameters}} \\ \hline
        learning rate & 0.00005  \\ \hline
        optimizer & AdamW  \\ \hline
        %scheduler & ReduceLROnPlateau  \\ \hline
        batch size & 512  \\ \hline
        epoch & 1000  \\ \hline
        $\alpha, \beta, \gamma, \delta, \epsilon$ & 0.1, 0.1, 0.2, 0.1, 0.2  \\ \hline
    \end{tabular}
    \end{center}
\end{table}

%%%%%%%%%%%%%%%%%%%%%%%%%%%%%%%%%%%%%%%%%%%%%%%%%%%%%%%%%%%%%%%%%%%%%%%%%%%%
%%%%%%%%%%%%%%%%%%%%%%%%%%%%%%%%%%%%%%%%%%%%%%%%%%%%%%%%%%%%%%%%%%%%%%%%%%%%
%%%%%%%%%%%%%%%%%%%%%%%%%%%%%%%%%%%%%%%%%%%%%%%%%%%%%%%%%%%%%%%%%%%%%%%%%%%%

\section{Detailed Explanation of Datasets and Experimental Setups}
\label{detailed_setup}

\subsection{Datasets}

\begin{table}[h]
\caption{Detailed information about the datasets.}
\label{datasets}
\begin{center}
\begin{tabular}{lllrrrl}
name&acronym&source&count&mean&std \\
\\ \hline \\
Abraham Descriptor S&AS&Ochem&1925&1.05&0.68\\
Boiling Point&BP&Pubchem&7139&198.99&108.88\\
Collision Cross Section&CCS&Pubchem&4006&205.06&57.84\\
Critical Temperature&CT&Ochem&242&626.04&120.96\\
Dielectric Constant&DK&Ochem&1007&0.80&0.41\\
Density&DS&Pubchem&3079&1.07&0.29\\
Enthalpy of Fusion&EF&Ochem&2188&1.32&0.32\\
Ionization Potential&IP&Pubchem&272&10.00&1.63\\
Kovats Retention Index&KRI&Pubchem&73507&2071.20&719.34\\
Log P&LP&Pubchem&28268&11.17&9.89\\
Polarizability&POL&CCCB&241&0.84&0.26\\
Surface Tension&ST&Pubchem&379&29.01&10.36\\
Viscosity&VS&Pubchem&294&0.47&0.87\\
Heat of Vaporization&HV&Pubchem&525&43.77&18.08
\end{tabular}
\end{center}
\end{table}

We utilized 14 different molecular property datasets sourced from three open-access databases, as detailed in Table~\ref{datasets} and the descriptions below, for the evaluation of GEAR. Prior to training, the datasets were carefully curated to remove entries with incorrectly specified units, typographical errors, or extreme measurement conditions. All datasets were normalized using their respective means and standard deviations to ensure consistency during training.

\begin{figure}[ht]
\begin{center}
\includegraphics[width=1\linewidth]{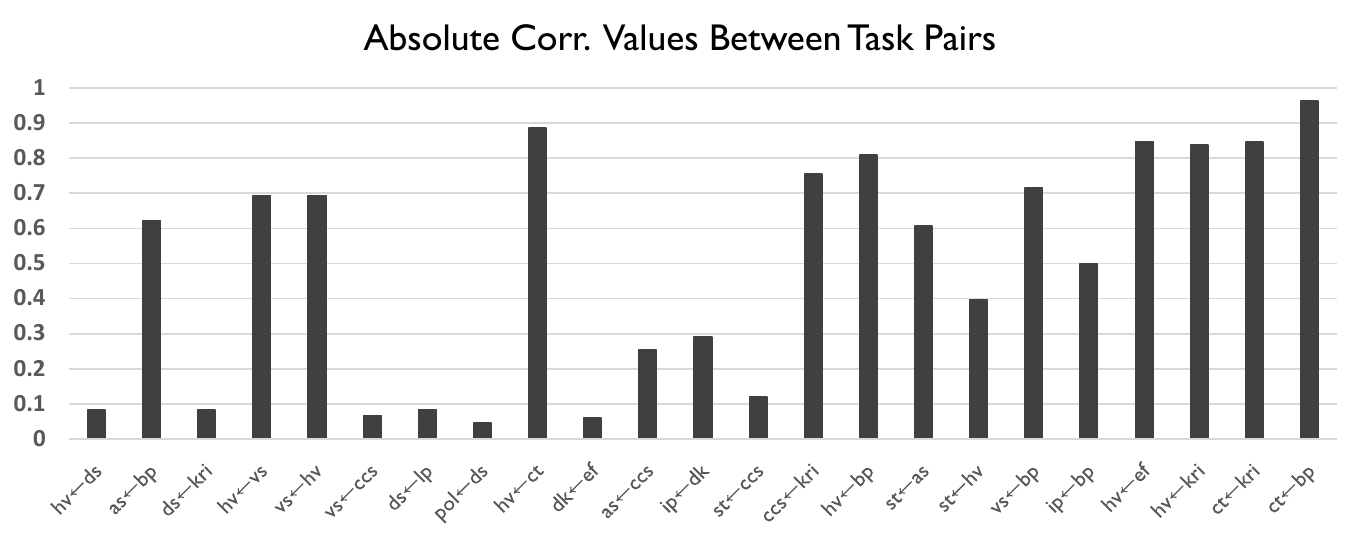}
\end{center}
\caption{Pearson correlation between overlapping data points in target dataset and source dataset.}
\label{fig:corr}
\end{figure}

From these datasets, we selected 23 source–target task pairs, considering the number of data points available in each dataset to maintain balance. Additionally, to ensure a fair and unbiased evaluation, we deliberately selected task pairs exhibiting a wide range of correlations, as illustrated in Figure~\ref{fig:corr}.

Finally, we provide an explicit description of the physical meaning associated with each dataset to facilitate better understanding and context.

\begin{itemize}
    \item {\bf AS} : The solute dipolarity/polarizability.
    \item {\bf BP} : The temperature at which this compound changes state from liquid to gas at a given atmospheric pressure.
    \item {\bf CCS} : The effective area for the interaction between an individual ion and the neutral gas through which it is traveling.
    \item {\bf CT} : The temparature when no gas can become liquid no matter how high the pressure is.
    \item {\bf DK} : The ratio of the electric permeability of the material to the electric permeability of free space.
    \item {\bf DS} : The mass of a unit volume of a compound.
    \item {\bf EF} : The change in enthalpy resulting from the addition or removal of heat from 1 mole of a substance to change its state from a solid to a liquid.
    \item {\bf IP} : The amount of energy required to remove an electron from an isolated atom or molecule.
    \item {\bf KRI} : The rate at which a compound is processed through a gas chromatography column.
    \item {\bf LP} : Logarithmic form of the ratio of concentrations of a compound in a mixture of octanol and water at equilibrium.
    \item {\bf POL} : The tendency of matter, when subjected to an electric field, to acquire an electric dipole moment in proportion to that applied field.
    \item {\bf ST} : The property of the surface of a liquid that allows it to resist an external force
    \item {\bf VS} : A measure of a fluid's resistance to flow.
    \item {\bf HV} : The quantity of heat that must be absorbed if a certain quantity of liquid is vaporized at a constant temperature.
\end{itemize}

\subsection{Experimental Setups}
For the evaluation of GEAR, we compared its performance against seven benchmark models: GATE, STL, MTL, KD, global structure-preserving loss-based KD (GSP-KD), and transfer learning (either retraining the entire model or only the head network). All baselines share the same base architecture, with minor adjustments specific to each method.

GATE shares nearly identical network parameters with GEAR for the encoder and head networks. However, for the transfer module, GEAR requires maintaining the input and output vector dimensions across each layer. Accordingly, the hidden dimensions were adjusted to [50, 50, 50], [50,50,50] instead of [100, 100, 100], [100,100,100]. Other hyperparameters strictly follow those introduced in the original paper~\citet{ko2023geometricallyalignedtransferencoder}.

In the MTL setup, the backbone and bottleneck layers are shared between the two tasks, while separate head networks are maintained for each task. For the KD baseline, latent vectors from the bottleneck are used as targets for knowledge distillation, with the distillation loss weighted at 0.1.

Graph Contrastive Representation Distillation (G-CRD) originally incorporates both contrastive and GSP losses~\citet{joshi2022representation}. However, since contrastive loss is unsuitable for regression tasks, we adopt only the GSP loss component. In GSP-KD, node features from the final layer of the backbone are used to compute pairwise distances, serving as the distillation targets. The loss ratio for GSP-based distillation is similarly set to 0.1.

Training is conducted for a maximum of 600 epochs, with the best model selected based on early stopping criteria.

%%%%%%%%%%%%%%%%%%%%%%%%%%%%%%%%%%%%%%%%%%%%%%%%%%%%%%%%%%%%%%%%%%%%%%%%%%%%
%%%%%%%%%%%%%%%%%%%%%%%%%%%%%%%%%%%%%%%%%%%%%%%%%%%%%%%%%%%%%%%%%%%%%%%%%%%%
%%%%%%%%%%%%%%%%%%%%%%%%%%%%%%%%%%%%%%%%%%%%%%%%%%%%%%%%%%%%%%%%%%%%%%%%%%%%

\section{Experimental Results} \label{bare_result}
We express explicit test results in this section. A total of 23 task pairs from 14 distinct datasets were thoroughly evaluated across eight different models. The full experimental results are presented across four tables. In each table, the best result for each task is highlighted with bold and underline, while the second-best result is underlined.

GEAR consistently outperforms other conventional methods by a significant margin. In both the random split and scaffold split settings, GEAR achieves the best performance on $73.91 \%$ of the tasks. Furthermore, when considering both first and second place rankings, GEAR ranks within the top two for $ 95.65 \% $ of all tasks.

\begin{table}[]
    \caption{Random Split Result (part 1)}
    \label{random_result_1}
    \begin{center}
    \begin{tabular}{ c  c  c  c  c  c  c  c  c }
    
         & \multicolumn{2}{c}{GEAR}  & \multicolumn{2}{c}{GATE}                    & \multicolumn{2}{c}{STL} & \multicolumn{2}{c}{MTL} \\
         \hline
       Tasks  & RMSE                  & STD                                                & RMSE                      & STD                 & RMSE                          & STD             & RMSE                          & STD            \\
         \hline
\bf{hv $\leftarrow$ ds} & \bf{\underline{0.8761}} & 0.1145 & \underline{0.9221} & 0.0612 & 0.9574 & 0.0519 & 0.9782 & 0.0782 \\
\bf{as $\leftarrow$ bp} & 0.4375 & 0.0188 & 0.4583 & 0.0193 & 0.5125 & 0.0085 & \underline{0.4370} & 0.0119 \\
\bf{ds $\leftarrow$ kri} & \bf{\underline{0.2796}} & 0.0492 & \underline{0.4145} & 0.0172 & 0.4154 & 0.0045 & 0.4172 & 0.0102 \\
\bf{hv $\leftarrow$ vs} & \bf{\underline{0.5711}} & 0.0358 & \underline{0.9116} & 0.0522 & 0.9574 & 0.0519 & 0.9700 & 0.1052 \\
\bf{vs $\leftarrow$ hv} & \bf{\underline{0.3364}} & 0.0513 & \underline{0.5471} & 0.0719 & 0.5947 & 0.0357 & 0.5535 & 0.0353 \\
\bf{st $\leftarrow$ as} & \bf{\underline{0.6045}} & 0.0981 & \underline{0.6689} & 0.0413 & 0.9902 & 0.0729 & 1.0272 & 0.0244 \\
\bf{ds $\leftarrow$ lp} & \bf{\underline{0.2677}} & 0.0567 & \underline{0.4046} & 0.0142 & 0.4154 & 0.0045 & 0.4133 & 0.0135 \\
\bf{pol $\leftarrow$ ds} & \underline{0.2820} & 0.0362 & 0.3431 & 0.0475 & 0.3460 & 0.0291 & 0.4367 & 0.1213 \\
\bf{vs $\leftarrow$ bp} & \bf{\underline{0.4299}} & 0.0771 & \underline{0.4457} & 0.0151 & 0.5947 & 0.0357 & 0.4516 & 0.0366 \\
\bf{dk $\leftarrow$ ef} & \bf{\underline{0.3748}} & 0.0092 & 0.4331 & 0.0140 & 0.4331 & 0.0358 & 0.4498 & 0.0126 \\
\bf{as $\leftarrow$ ccs} & \bf{\underline{0.4400}} & 0.0136 & \underline{0.4648} & 0.0139 & 0.5125 & 0.0085 & 0.4677 & 0.0220 \\
\bf{ct $\leftarrow$ bp} & \bf{\underline{0.1481}} & 0.0138 & 0.1742 & 0.0034 & 0.2549 & 0.1247 & 0.1707 & 0.0132 \\
\bf{st $\leftarrow$ ccs} & \bf{\underline{0.9222}} & 0.0232 & \underline{0.9546} & 0.0452 & 0.9902 & 0.0729 & 1.0361 & 0.0737 \\
\bf{ccs $\leftarrow$ kri} & \underline{0.2426} & 0.0108 & 0.2476 & 0.0034 & 0.2936 & 0.0110 & 0.2524 & 0.0042 \\
\bf{hv $\leftarrow$ bp} & \bf{\underline{0.6252}} & 0.0320 & \underline{0.7251} & 0.0581 & 0.9574 & 0.0519 & 0.7550 & 0.0432 \\
\bf{vs $\leftarrow$ ccs} & \bf{\underline{0.3364}} & 0.0513 & 0.5233 & 0.0323 & 0.5947 & 0.0357 & 0.5792 & 0.0228 \\
\bf{st $\leftarrow$ hv} & \bf{\underline{0.5443}} & 0.0841 & 0.7647 & 0.0622 & 0.9902 & 0.0729 & \underline{0.7179} & 0.0259 \\
\bf{hv $\leftarrow$ ct} & \bf{\underline{0.7481}} & 0.1196 & 0.9399 & 0.0896 & 0.9574 & 0.0519 & 1.1118 & 0.1633 \\
\bf{ip $\leftarrow$ bp} & \bf{\underline{0.4363}} & 0.0307 & 0.5476 & 0.0642 & 0.6695 & 0.0660 & 0.6067 & 0.0345 \\
\bf{hv $\leftarrow$ ef} & \underline{0.7409} & 0.1171 & \bf{\underline{0.6131}} & 0.0966 & 0.9574 & 0.0519 & 0.8296 & 0.0999 \\
\bf{hv $\leftarrow$ kri} & \underline{0.6990} & 0.0888 & \bf{\underline{0.5410}} & 0.0732 & 0.9574 & 0.0519 & 0.8631 & 0.0354 \\
\bf{ct $\leftarrow$ kri} & \bf{\underline{0.1481}} & 0.0138 & \underline{0.1658} & 0.0136 & 0.2549 & 0.1247 & 0.1716 & 0.0090 \\
\bf{ip $\leftarrow$ dk} & \bf{\underline{0.5159}} & 0.0362 & 0.6510 & 0.0381 & 0.6695 & 0.0660 & 0.7083 & 0.0226 \\
\hline
mean    &  \bf{\underline{0.4785}} & 0.0514 & \underline{0.5592} & 0.0412 & 0.6642 & 0.0487 & 0.6263 & 0.0443 \\
\hline
         & Count                 & Ratio                                              & Count                     & Ratio               & Count                         & Ratio           & Count                         & Ratio          \\
         \hline
1st     & 18 & 78.26 \% & 2 & 8.70 \% & 0 & 0 \% & 0 & 0 \% \\
\hline
2nd     &22 & 95.65 \% & 13 & 56.52 \% & 0 & 0 \% & 2 & 8.70 \% \\
\hline
\end{tabular}
\end{center}
\end{table}

\begin{table}[]
    \caption{Random Split Result (part 2)}
    \label{random_result_2}
    \begin{center}
    \begin{tabular}{ c  c  c  c  c  c  c  c  c }
    
        & \multicolumn{2}{c}{KD}  & \multicolumn{2}{c}{GSP-KD}  & \multicolumn{2}{c}{Transfer All}  & \multicolumn{2}{c}{Transfer Head}  \\
        \hline
        Tasks & RMSE                            & STD & RMSE                            & STD              & RMSE                                   & STD                     & RMSE                            & STD                             \\
        \hline
\bf{hv ← ds} & 1.3726 & 0.2930 & 0.9321 & 0.0487 & 1.0428 & 0.1165 & 1.1166 & 0.0024 \\
\bf{as ← bp} & 0.5426 & 0.0335 & 0.5315 & 0.0151 & \bf{\underline{0.4325}} & 0.0104 & 0.7712 & 0.0105 \\
\bf{ds ← kri} & 0.4403 & 0.0119 & 0.4147 & 0.0063 & 0.4414 & 0.0154 & 0.8842 & 0.0049 \\
\bf{hv ← vs} & 1.1995 & 0.1419 & 0.9154 & 0.0130 & 0.9937 & 0.0821 & 1.0091 & 0.0181 \\
\bf{vs ← hv} & 0.5878 & 0.0264 & 0.5619 & 0.0223 & 0.5712 & 0.0232 & 0.7215 & 0.0392 \\
\bf{st ← as} & 1.1601 & 0.0396 & 0.9938 & 0.0141 & 1.1296 & 0.1302 & 1.0045 & 0.0220 \\
\bf{ds ← lp} & 0.4378 & 0.0086 & 0.4106 & 0.0077 & 0.4280 & 0.0136 & 0.9111 & 0.0022 \\
\bf{pol ← ds} & 0.3089 & 0.0270 & \bf{\underline{0.2603}} & 0.0270 & 0.3741 & 0.0303 & 0.9060 & 0.0141 \\
\bf{vs ← bp} & 0.6076 & 0.0241 & 0.5932 & 0.0097 & 0.5445 & 0.0239 & 0.7220 & 0.0645 \\
\bf{dk ← ef} & \underline{0.3852} & 0.0238 & 0.4230 & 0.0133 & 0.3936 & 0.0164 & 0.9380 & 0.0026 \\
\bf{as ← ccs} & 0.5364 & 0.0211 & 0.5457 & 0.0150 & 0.4741 & 0.0148 & 0.9935 & 0.0033 \\
\bf{ct ← bp} & 0.1690 & 0.0079 & 0.2018 & 0.0093 & \underline{0.1563} & 0.0044 & 0.6847 & 0.0186 \\
\bf{st ← ccs} & 1.1731 & 0.0730 & 0.9595 & 0.0405 & 1.1334 & 0.0687 & 1.1039 & 0.0046 \\
\bf{ccs ← kri} & 0.2622 & 0.0117 & 0.2698 & 0.0095 & \bf{\underline{0.2273}} & 0.0016 & 0.6166 & 0.0567 \\
\bf{hv ← bp} & 1.1983 & 0.1815 & 0.9051 & 0.0571 & 0.8267 & 0.0417 & 0.8829 & 0.0499 \\
\bf{vs ← ccs} & 0.6027 & 0.0127 & 0.5269 & 0.0167 & \underline{0.4868} & 0.0119 & 0.8684 & 0.0116 \\
\bf{st ← hv} & 1.1270 & 0.0184 & 0.9618 & 0.0086 & 1.0290 & 0.0945 & 1.0102 & 0.0138 \\
\bf{hv ← ct} & 1.5114 & 0.1845 & \underline{0.9207} & 0.0112 & 1.2072 & 0.0460 & 1.0302 & 0.0186 \\
\bf{ip ← bp} & 0.5624 & 0.0273 & \underline{0.4631} & 0.0037 & 0.9816 & 0.2334 & 0.8732 & 0.0293 \\
\bf{hv ← ef} & 1.3659 & 0.2587 & 0.8112 & 0.0463 & 1.0818 & 0.1021 & 0.9616 & 0.0478 \\
\bf{hv ← kri} & 1.3739 & 0.2487 & 0.9191 & 0.0676 & 0.9080 & 0.0510 & 1.0715 & 0.0145 \\
\bf{ct ← kri} & 0.1586 & 0.0102 & 0.2080 & 0.0057 & 0.1661 & 0.0075 & 0.8349 & 0.0279 \\
\bf{ip ← dk} & 0.5508 & 0.0100 & \underline{0.5257} & 0.0192 & 0.6099 & 0.0273 & 1.0336 & 0.0085 \\
mean & 0.7667 & 0.0737 & 0.6198 & 0.0212 & 0.6800 & 0.0507 & 0.9108 & 0.0211 \\
\hline
mean & 0.7667 & 0.0737 & 0.6198 & 0.0212 & 0.6800 & 0.0507 & 0.9108 & 0.0211 \\
\hline
        & Count & Ratio & Count                           & Ratio            & Count                                  & Ratio                   & Count                           & Ratio                           \\
        \hline
1st     & 0 & 0 \% & 1 & 4.35 \% & 2 & 8.70 \% & 0 & 0 \% \\
\hline
2nd     & 1 & 4.35 \% & 4 & 17.39 \% & 4 & 17.39 \% & 0 & 0 \% \\
\hline
\end{tabular}
\end{center}
\end{table}

\begin{table}[]
    \caption{Scaffold Split Result (part 1)}
    \label{scaffold_part1}
    \begin{center}
    \begin{tabular}{c c c c c c c c c}
        & \multicolumn{2}{c}{GEAR} & \multicolumn{2}{c}{GATE} & \multicolumn{2}{c}{STL} & \multicolumn{2}{c}{MTL} \\
        \hline
        Tasks & RMSE & STD & RMSE & STD & RMSE & STD & RMSE & STD \\
        \hline
\bf{hv ← ds} & \underline{0.6101} & 0.0210 & 0.6939 & 0.0996 & 0.6744 & 0.1079 & 0.6465 & 0.0776 \\
\bf{as ← bp} & \bf{\underline{1.0016}} & 0.0073 & \underline{1.0495} & 0.0256 & 1.2828 & 0.0724 & 1.1677 & 0.1068 \\
\bf{ds ← kri} & \bf{\underline{0.4261}} & 0.0017 & \underline{0.4395} & 0.0108 & 0.4477 & 0.0052 & 0.4849 & 0.0061 \\
\bf{hv ← vs} & \bf{\underline{0.5731}} & 0.0470 & 0.7174 & 0.0796 & 0.6744 & 0.1079 & 0.9954 & 0.2059 \\
\bf{vs ← hv} & \underline{0.6323} & 0.0441 & \bf{\underline{0.6120}} & 0.0639 & 0.9816 & 0.1267 & 0.8535 & 0.0558 \\
\bf{st ← as} & \bf{\underline{0.6980}} & 0.0832 & \underline{0.7540} & 0.0660 & 0.8041 & 0.1062 & 1.0254 & 0.0251 \\
\bf{ds ← lp} & \underline{0.4236} & 0.0036 & \bf{\underline{0.4049}} & 0.0102 & 0.4477 & 0.0052 & 0.4517 & 0.0184 \\
\bf{pol ← ds} & 0.9902 & 0.0697 & \underline{0.9040} & 0.0852 & 0.9604 & 0.1056 & 1.4198 & 0.0796 \\
\bf{vs ← bp} & \bf{\underline{0.5242}} & 0.0418 & 0.6121 & 0.0297 & 0.9816 & 0.1267 & \underline{0.5686} & 0.0276 \\
\bf{dk ← ef} & \bf{\underline{0.5229}} & 0.0166 & 0.7122 & 0.0545 & 0.7028 & 0.0391 & 0.6549 & 0.0210 \\
\bf{as ← ccs} & \bf{\underline{1.0016}} & 0.0073 & 1.1313 & 0.0496 & 1.2828 & 0.0724 & \underline{1.1197} & 0.0558 \\
\bf{ct ← bp} & \bf{\underline{0.3275}} & 0.0329 & \underline{0.3883} & 0.0203 & 1.4436 & 0.1150 & 0.4359 & 0.0126 \\
\bf{st ← ccs} & \bf{\underline{0.6975}} & 0.0833 & \underline{0.7281} & 0.0586 & 0.8041 & 0.1062 & 0.9905 & 0.0737 \\
\bf{ccs ← kri} & \bf{\underline{0.5111}} & 0.0044 & \underline{0.5292} & 0.0094 & 0.5489 & 0.0107 & 0.5297 & 0.0083 \\
\bf{hv ← bp} & \underline{0.4671} & 0.0136 & 0.4821 & 0.0132 & 0.6744 & 0.1079 & \bf{\underline{0.4668}} & 0.0169 \\
\bf{vs ← ccs} & \bf{\underline{0.5611}} & 0.0676 & \underline{0.6126} & 0.0671 & 0.9816 & 0.1267 & 0.8186 & 0.0790 \\
\bf{st ← hv} & \bf{\underline{0.6980}} & 0.0832 & \underline{0.7209} & 0.0412 & 0.8041 & 0.1062 & 0.7237 & 0.0276 \\
\bf{hv ← ct} & \bf{\underline{0.5038}} & 0.0236 & 0.6579 & 0.0678 & 0.6744 & 0.1079 & 0.6633 & 0.0660 \\
\bf{ip ← bp} & \bf{\underline{0.4064}} & 0.0300 & 0.4668 & 0.0179 & 0.5780 & 0.1475 & 0.5540 & 0.0587 \\
\bf{hv ← ef} & \bf{\underline{0.5038}} & 0.0236 & 0.6406 & 0.0335 & 0.6744 & 0.1079 & 0.7879 & 0.0643 \\
\bf{hv ← kri} & \bf{\underline{0.4812}} & 0.0150 & \underline{0.5084} & 0.0264 & 0.6744 & 0.1079 & 0.6204 & 0.0269 \\
\bf{ct ← kri} & \underline{0.4256} & 0.0214 & \bf{\underline{0.3902}} & 0.0140 & 1.4436 & 0.1150 & 0.5173 & 0.0927 \\
\bf{ip ← dk} & \bf{\underline{0.3984}} & 0.0232 & \underline{0.4335} & 0.0119 & 0.5780 & 0.1475 & 0.5335 & 0.1016 \\
\hline
mean & \bf{\underline{0.5820}} & 0.0333 & \underline{0.6343} & 0.0416 & 0.8313 & 0.0949 & 0.7404 & 0.0569 \\
\hline
        & Count & Ratio & Count                           & Ratio            & Count                                  & Ratio                   & Count                           & Ratio                           \\
        \hline
1st     & 17 & 73.91 \% & 3 & 13.04 \% & 0 & 0 \% & 1 & 4.35 \% \\
\hline
2nd     & 22 & 95.65 \% & 14 & 60.87 \% & 0 & 0 \% & 3 & 13.04 \% \\
\hline
\end{tabular}
\end{center}
\end{table}

\begin{table}[]
    \caption{Scaffold Split Result (part 2)}
    \label{scaffold_part2}
    \begin{center}
    \begin{tabular}{c c c c c c c c c}
        & \multicolumn{2}{c}{KD} & \multicolumn{2}{c}{GSP-KD} & \multicolumn{2}{c}{Transfer All} & \multicolumn{2}{c}{Transfer Head} \\
        \hline
        Tasks & RMSE & STD & RMSE & STD & RMSE & STD & RMSE & STD \\
        \hline
\bf{hv ← ds} & \bf{\underline{0.5920}} & 0.0466 & 0.7606 & 0.0810 & 0.8659 & 0.0788 & 0.9584 & 0.0339 \\
\bf{as ← bp} & 1.3580 & 0.0136 & 1.2340 & 0.0294 & 1.1478 & 0.0264 & 1.0935 & 0.0079 \\
\bf{ds ← kri} & 0.5409 & 0.0480 & 0.4467 & 0.0104 & 0.8753 & 0.1134 & 1.0928 & 0.0482 \\
\bf{hv ← vs} & 0.8948 & 0.2294 & \underline{0.6536} & 0.0345 & 0.7520 & 0.1666 & 0.7924 & 0.0595 \\
\bf{vs ← hv} & 1.2597 & 0.3638 & 0.6377 & 0.0253 & 0.9217 & 0.1575 & 0.9179 & 0.0539 \\
\bf{st ← as} & 1.7083 & 0.1608 & 0.9335 & 0.0954 & 1.2604 & 0.0946 & 1.0780 & 0.0613 \\
\bf{ds ← lp} & 0.5221 & 0.0328 & 0.4685 & 0.0111 & 0.4664 & 0.0121 & 1.0410 & 0.0026 \\
\bf{pol ← ds} & 1.3309 & 0.1998 & \bf{\underline{0.8475}} & 0.0627 & 1.0385 & 0.2146 & 1.3204 & 0.0491 \\
\bf{vs ← bp} & 0.9371 & 0.2386 & 0.6599 & 0.0204 & 1.1532 & 0.1766 & 1.0135 & 0.0820 \\
\bf{dk ← ef} & 0.8189 & 0.0462 & \underline{0.6353} & 0.0171 & 0.7417 & 0.0384 & 0.7963 & 0.0071 \\
\bf{as ← ccs} & 1.3773 & 0.0781 & 1.1272 & 0.0778 & 1.2925 & 0.0606 & 1.4530 & 0.0143 \\
\bf{ct ← bp} & 1.2459 & 0.1199 & 1.1837 & 0.0586 & 0.5644 & 0.0530 & 0.9347 & 0.0316 \\
\bf{st ← ccs} & 1.5402 & 0.1418 & 0.7344 & 0.0187 & 0.9075 & 0.0431 & 1.2596 & 0.0287 \\
\bf{ccs ← kri} & 0.5534 & 0.0190 & 0.5356 & 0.0115 & 0.5640 & 0.0137 & 0.7904 & 0.0159 \\
\bf{hv ← bp} & 0.6271 & 0.0868 & 0.7403 & 0.0889 & 0.6093 & 0.0422 & 0.8111 & 0.0251 \\
\bf{vs ← ccs} & 1.3034 & 0.5354 & 0.8027 & 0.0159 & 0.7271 & 0.0828 & 1.2282 & 0.0243 \\
\bf{st ← hv} & 1.5256 & 0.1906 & 0.7417 & 0.0206 & 1.4243 & 0.0627 & 1.0047 & 0.0813 \\
\bf{hv ← ct} & 0.7925 & 0.2694 & \underline{0.6428} & 0.0080 & 0.9499 & 0.2579 & 0.8089 & 0.0532 \\
\bf{ip ← bp} & \underline{0.4205} & 0.0240 & 0.4579 & 0.0207 & 0.4419 & 0.0371 & 0.9704 & 0.0399 \\
\bf{hv ← ef} & 0.6773 & 0.1553 & \underline{0.5862} & 0.0375 & 1.0003 & 0.1719 & 0.9503 & 0.0307 \\
\bf{hv ← kri} & 0.6710 & 0.1524 & 0.5509 & 0.0252 & 0.6560 & 0.0408 & 0.9998 & 0.0311 \\
\bf{ct ← kri} & 1.3392 & 0.1076 & 1.2358 & 0.0373 & 1.1124 & 0.1265 & 1.2769 & 0.0193 \\
\bf{ip ← dk} & 0.4975 & 0.0769 & 0.4376 & 0.0255 & 0.5248 & 0.0471 & 1.0165 & 0.0521 \\
\hline
mean & 0.9797 & 0.1451 & 0.7415 & 0.0362 & 0.8694 & 0.0921 & 1.0265 & 0.0217 \\
\hline
        & Count & Ratio & Count                           & Ratio            & Count                                  & Ratio                   & Count                           & Ratio                           \\
        \hline
1st     & 1 & 4.35 \% & 1 & 4.35 \% & 0 & 0 \% & 0 & 0 \% \\
\hline
2nd     & 2 & 8.70 \% & 5 & 21.74 \% & 0 & 0 \% & 0 & 0 \% \\
\hline
\end{tabular}
\end{center}
\end{table}

%%%%%%%%%%%%%%%%%%%%%%%%%%%%%%%%%%%%%%%%%%%%%%%%%%%%%%%%%%%

\end{document}